\documentclass[lettersize,journal,hidelinks]{IEEEtran}

\usepackage{amsmath,amsfonts}
\usepackage{newtxmath}
\usepackage{graphicx}
\usepackage{tabularray}
\usepackage{subcaption,booktabs}
\usepackage{float} 
\usepackage{stfloats} 
\usepackage{shortcuts}
\usepackage{makecell}
\usepackage{multirow}
\usepackage{algorithm}
\usepackage{algpseudocode}
\usepackage{comment}
\usepackage[T1]{fontenc} % For proper encoding
\usepackage{orcidlink}

\newtheorem{definition}{Definition}

\begin{document}

\title{Memorization in deep learning: A survey}

\author{Jiaheng Wei\orcidlink{0009-0003-7180-4268},
Yanjun Zhang\orcidlink{0000-0001-5611-3483},
Leo Yu Zhang\orcidlink{0000-0001-9330-2662},~\IEEEmembership{Member,~IEEE,}
Ming Ding\orcidlink{0000-0002-3690-0321},~\IEEEmembership{Senior Member,~IEEE,}
Chao Chen\orcidlink{0000-0003-1355-3870},~\IEEEmembership{Member,~IEEE,} 
Kok-Leong Ong\orcidlink{0000-0003-4688-7674
},
Jun Zhang\orcidlink{0000-0002-2189-7801},~\IEEEmembership{Senior Member,~IEEE,} 
Yang Xiang\orcidlink{0000-0001-5252-0831},~\IEEEmembership{Fellow,~IEEE.}

\thanks{Manuscript received June 7, 2024.}

\thanks{Jiaheng Wei, Chao Chen, and Kok-Leong Ong are with the School of Accounting, Information System and Supply Chain, RMIT University, Melbourne, VIC 3001, Australia (e-mail: s3986349@student.rmit.edu.au; chao.chen@rmit.edu.au; kok-leong.ong2@rmit.edu.au).}

\thanks{Yanjun Zhang is with the School of Computer Science, University
of Technology Sydney, Sydney, NSW 2007, Australia (e-mail: Yanjun.Zhang@uts.edu.au).}

\thanks{Leo Yu Zhang is with the School of Information and Communication Technology, Griffith University, Brisbane, QLD 4215, Australia (e-mail: leo.zhang@griffith.edu.au).}

\thanks{Ming Ding is with Data61, CSIRO, Sydney, NSW 2015, Australia (e-mail: ming.ding@data61.csiro.au).}

\thanks{Jun Zhang and Yang Xiang are with the School of Science, Computing and Engineering Technologies, Swinburne University of Technology, Melbourne, VIC 3122, Australia (e-mail: junzhang@swin.edu.au; yxiang@swin.edu.au).}}

\markboth{Journal of \LaTeX\ Class Files,~Vol.~18, No.~9, September~2020}%
{How to Use the IEEEtran \LaTeX \ Templates}

\maketitle

\begin{abstract}
%This document describes the most common article elements and how to use the IEEEtran class with \LaTeX \ to produce files that are suitable for submission to the Institute of Electrical and Electronics Engineers (IEEE).  IEEEtran can produce conference, journal and technical note (correspondence) papers with a suitable choice of class options.

Deep Learning (DL) powered by Deep Neural Networks (DNNs) has revolutionized various domains, 
yet understanding the intricacies of DNN decision-making and learning processes remains a significant challenge. 
Recent investigations have uncovered an interesting memorization phenomenon in which DNNs tend to memorize specific details from examples rather than learning general patterns, 
affecting model generalization, security, and privacy. 
This raises critical questions about the nature of generalization in DNNs and their susceptibility to security breaches. 
In this survey, 
we present a systematic framework to organize memorization definitions based on the generalization and security/privacy domains and summarize memorization evaluation methods at both the example and model levels.
Through a comprehensive literature review, 
we explore DNN memorization behaviors and their impacts on security and privacy. 
% Memorization has some important properties, 
% including the ability to enhance generalization by capturing rare and atypical examples. 
%and overfitting is not responsible for memorization. 
We also introduce privacy vulnerabilities caused by memorization 
and the phenomenon of forgetting and explore its connection with memorization.
Furthermore, 
we spotlight various applications leveraging memorization and forgetting mechanisms, including noisy label learning, privacy preservation, and model enhancement. 
This survey offers the first-in-kind understanding of memorization in DNNs, 
providing insights into its challenges and opportunities for enhancing AI development while addressing critical ethical concerns.

% In this survey article, 
% we investigate the memorization phenomenon in Deep Neural Networks (DNNs), 
% where models learn specific details from examples rather than general patterns, 
% affecting model generalization, security, and privacy. 
% Through an extensive literature review, 
% we propose a framework categorizing memorization research into its impacts on standard training behaviours and security/privacy risks, 
% and examine memorization at both the example and model levels. 
% It also explores the related concept of forgetting and its applications. 
% Our key contributions include organizing memorization definitions,
% reviewing relevant studies, 
% and discussing the implications of memorization on AI technologies. 
% This comprehensive analysis aims to deepen the understanding of memorization in DNNs, 
% offering insights into its challenges and opportunities for enhancing AI development while addressing ethical concerns.

\end{abstract}

\section{Introduction}
In the development of artificial intelligence (AI), deep learning (DL) has emerged as an effective solution for various complex tasks like text generation~\cite{achiam2023gpt}, speech translation~\cite{dhanjal2024comprehensive}, etc. 
Deep neural network (DNN) as the main model architecture has been widely used in numerous innovative applications such as autonomous vehicles~\cite{tian2018deeptest, Hu_2023_CVPR, hu2023planning} and medical diagnosis~\cite{li2023dynamic, tu2024towards}. 

\begin{figure}[!htbp]
	\centering
	\subfloat[Memorization Phenomenon]
{\centering
\includegraphics[width=0.48\textwidth]{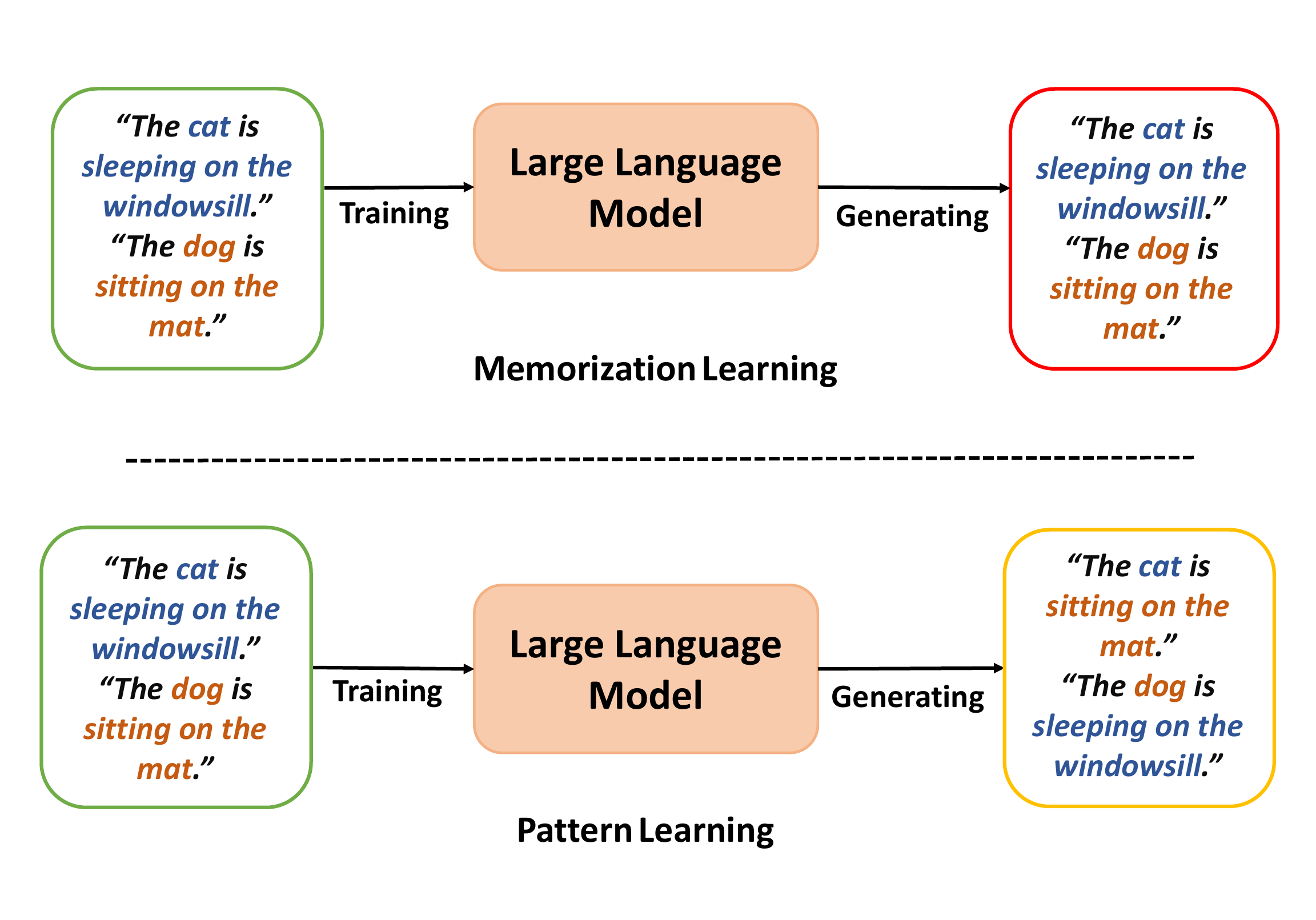}%
		\label{subfig:StoryFigure}}
	\hfil
	\subfloat[Memorization Level]{\includegraphics[width=0.48\textwidth]{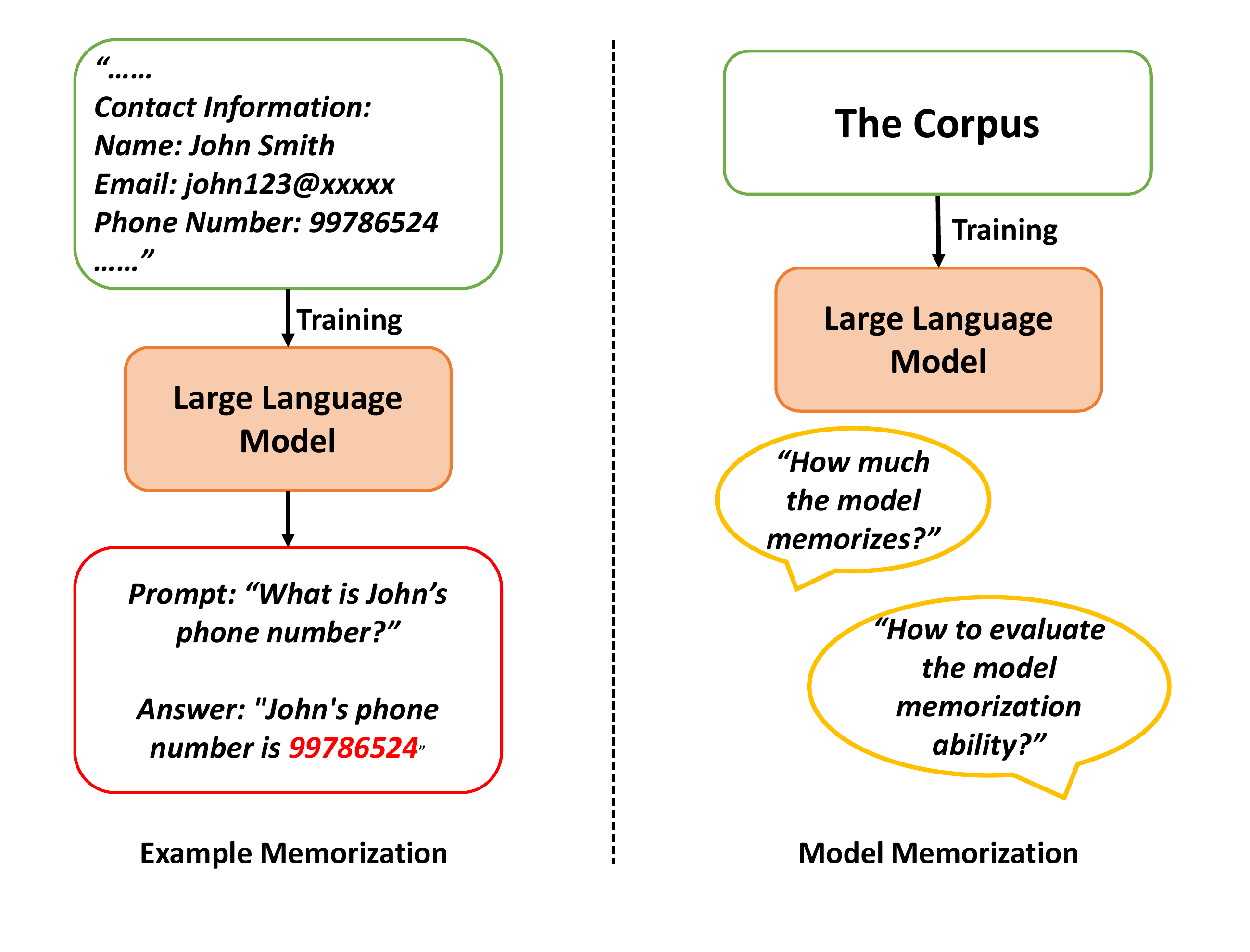}%
        \label{subfig:MemComponents}}
	\caption{The Direct Memorization Effect. In (a), we use an image generator to describe memorization. The upper part demonstrates the memorization effect and the lower part represents the common generation. For (b), the memorization effect has two different levels: Example Memorization and Model Memorization.}
\end{figure}
%\vspace{-3em}

However, it is still challenging to understand how DNNs make decisions and what they learn from the training data. Though researchers believe DNNs can learn patterns in the training data to achieve success in assigned tasks, a recent study found that DNNs are able to memorize the entire randomly labeled training dataset~\cite{zhang_UnderstandingDeepLearning_2017}, which illustrates that properties of the model family, or the regularization techniques fail to explain why large neural networks generalize well. DNNs may memorize particular features from training data instead of learning patterns to perform specific tasks. This attracts the community to explore the memorization mechanism and prompts researchers to rethink the generalization in DNNs. Additionally, this memorization phenomenon raises concerns about the security of AI because of potential privacy leakage risks and vulnerability against malicious attacks. Furthermore, the training dataset collected from the real world may contain significant noise and bias, and memorized data in DNNs may keep the noise and bias, impairing the usability and fairness of the models. 

So far, 
numerous papers have found the memorization effects that neural networks may memorize some training data in training with gradient descent~\cite{carlini_ExtractingTrainingData_2021,carlini_ExtractingTrainingData_2023,zhang_UnderstandingDeepLearning_2017,feldman_WhatNeuralNetworks_2020,zheng_EmpiricalStudyMemorization_2022}. Current memorization studies mainly focus on two domains: 
the behaviors in standard training and the security/privacy risks. We summarize explicit memorization definitions in literature based on generalization and security/privacy.
However, there is a lack of a widely adopted definition for memorization, making describing and discussing the memorization concept challenging.
Many relevant works provide inconsistent, sometimes contradictory, definitions of memorization. Especially, many works directly apply the word "memorization" as the synonymous words of "learning" and "fitting".
Thus, we adopt the following terms for facilitating discussion:
\textbf{Memorization Learning} refers to DNNs learning specific details or particular features of examples, 
while common \textbf{Pattern Learning} indicates DNNs learning the common patterns or generalized features of the data distribution.In Figure~\ref{subfig:StoryFigure}, 
we use a large language model to illustrate memorization learning and pattern learning.
We use the word "generalization" to define the model performance on the new, unseen data. Suppose there is no extra explanation, all terms like "memorization", "memorization effect", and "memorization phenomenon" point to memorization learning. Moreover, we think pattern learning and memorization learning together constitute the learning path of DNNs.

\begin{figure*}[!htbp]
    \centering
    \includegraphics[width=1\textwidth]{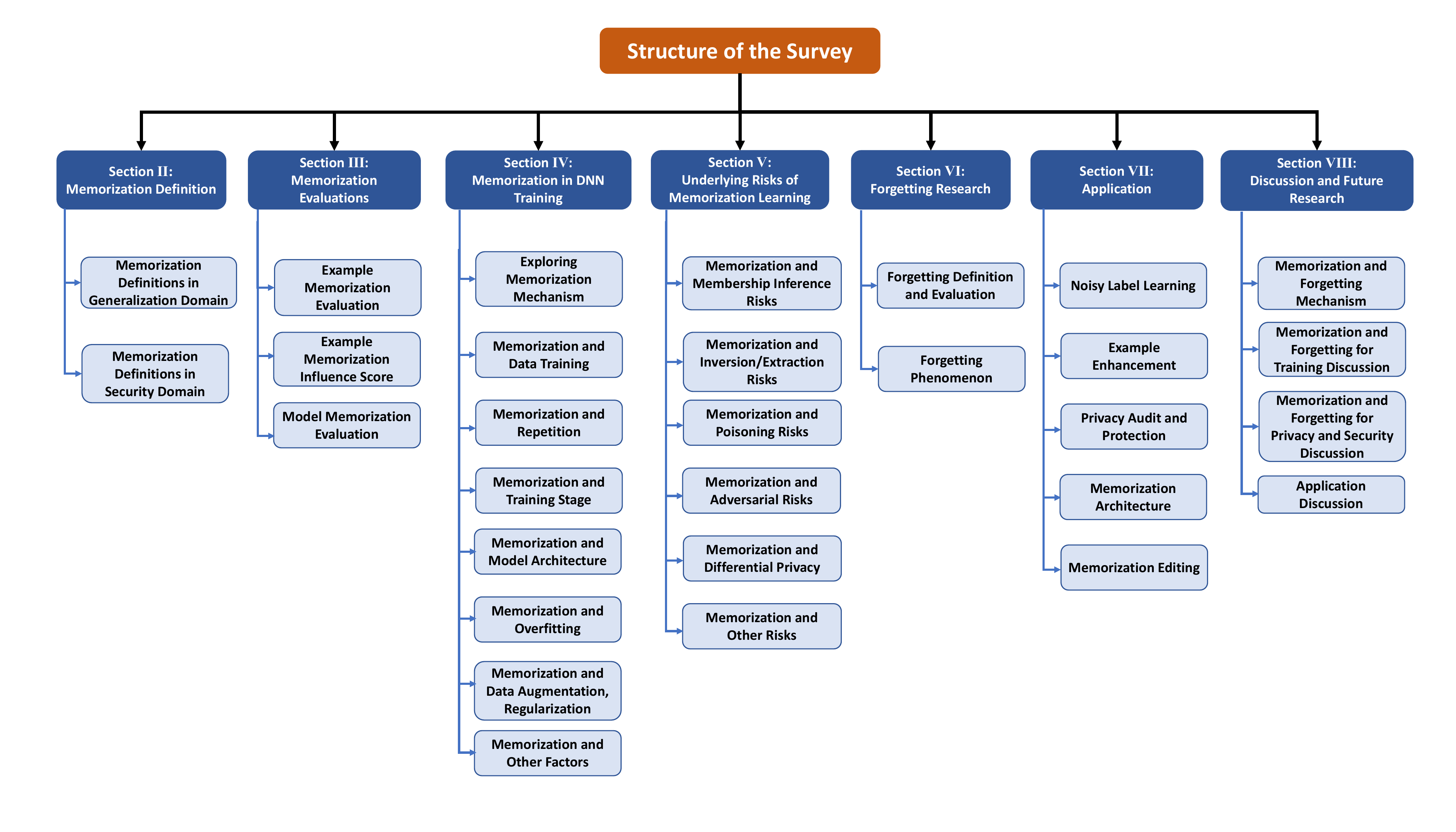}
    \caption{Paper Structure.}
    \label{fig:paperStructure}
\end{figure*}

Moreover, 
memorization is a complex concept that requires us to consider it at various levels. 
In our opinion, 
memorization learning and pattern learning operate at a feature level. 
However, 
understanding the features of neural networks directly is exceedingly difficult for humans. 
Hence, 
we mainly study memorization at the example level and model level as illustrated in Figure ~\ref{subfig:MemComponents}. 

Intuitively, example memorization and model memorization indicate the objects of study are examples and models. Consequently, 
memorization concepts at different levels inspire distinct memorization evaluation methods. Example memorization evaluation tries to ensure if an example is memorized including differential evaluation and probabilistic evaluation. On the other hand, model memorization evaluation measures how much models memorize or the memorization ability of models. We summarize various approaches to three main methods: noisy label evaluation, recurrence evaluation, and extraction evaluation.

After the definitions and evaluation methods, we systematically review related literature. For memorization behaviors in standard training, existing studies investigate the relationships between the memorization effect and training data, training stages, model architecture, overfitting, regularization, and other factors. One study~\cite{feldman_DoesLearningRequire_2020,feldman_WhatNeuralNetworks_2020} provides an interesting conclusion that memorization learning improves the generalization of models because the memorization of rare and atypical examples actually contributes to the generalization performance of similar rare subgroups, which is adverse to some early opinions. Additionally, some evidence~\cite{carlini_SecretSharerEvaluating_2019,maini_CanNeuralNetwork_2023,tirumala_MemorizationOverfittingAnalyzing_2022} shows overfitting is not responsible for memorization. Memorization is a persistent process in training. For security/privacy risks, 
the memorized particular features become multiple risk sources like membership inference risks and extraction risks, 
enabling attackers to exploit the memorization mechanism to invade privacy and violate the security rules of DNNs. In contrast, some risks like adversary attack risks are not obviously related to the memorization mechanism.

On a related aspect, the forgetting phenomenon is closely connected to the memorization effect. 
Thus, we also discuss and review the forgetting effect. We explore useful forgetting definitions and evaluation methods and summarize relevant forgetting phenomenon studies. 

Additionally, we also review numerous applications utilizing the memorization and forgetting mechanisms. These applications like noisy label learning, example enhancement, privacy audit and protection, memorization architecture, and model editing, take advantage of different properties of memorization.

In summary, we attempt to organize the memorization definitions and evaluation methods and review relevant literature, aiming to build a scientific and effective framework and help the readers understand the memorization mechanism and its influence on model training and system security.
Additionally, we also explore the forgetting phenomenon and illustrate some potential applications of the memorization and forgetting mechanisms. 
We hope this survey can help the research community have a general understanding of the memorization phenomenon. 
The key contributions of this survey are summarized as follows:
\begin{itemize}
    \item \textbf{Organizing definitions.} We propose a framework to organize all existing memorization definitions and evaluation methods. We also explain the scope and limitations of these definitions and evaluation methods. 
    \item \textbf{Comprehensive review.} We review relevant memorization studies on its behaviors in the standard training and security/privacy risks. Moreover, we also investigate its connection with the forgetting studies and some possible applications.
    \item \textbf{Discussion.} In this survey, we thoroughly discuss the memorization mechanism and how memorization effects can boost other relevant technologies.
\end{itemize}

%Add Ref
The survey is organized into the following sections, each focusing on a different aspect of memorization in deep learning as we present in Figure~\ref{fig:paperStructure}. Section~\ref{sec:definition} provides existing memorization definitions and Section~\ref{sec:eval} lists the memorization evaluation methods based on various levels. Section~\ref{sec:memorizationBehaviors} delves into the memorization behaviors, presenting how memorization affects each training component and its relationship with overfitting, data augmentation, and regularization technology. Section~\ref{sec:securityAndPrivacy} presents a review of memorization-associated risks that memorized particular features enhance privacy risks. Section~\ref{sec:forgetting} explores the forgetting phenomenon, which is the opposite of memorization. Section~\ref{sec:application} demonstrates the underlying application of the memorization effects including the noisy label learning, example enhancement technology, privacy audit and protection, and memorization architecture. Section~\ref{sec:discussion} comprehensively discusses the memorization phenomenon's influence on standard training and security/privacy risks and how it enlightens and explains other technologies or phenomena.

\begin{figure*}[!htbp]
    \centering
    \includegraphics[width=1\textwidth]{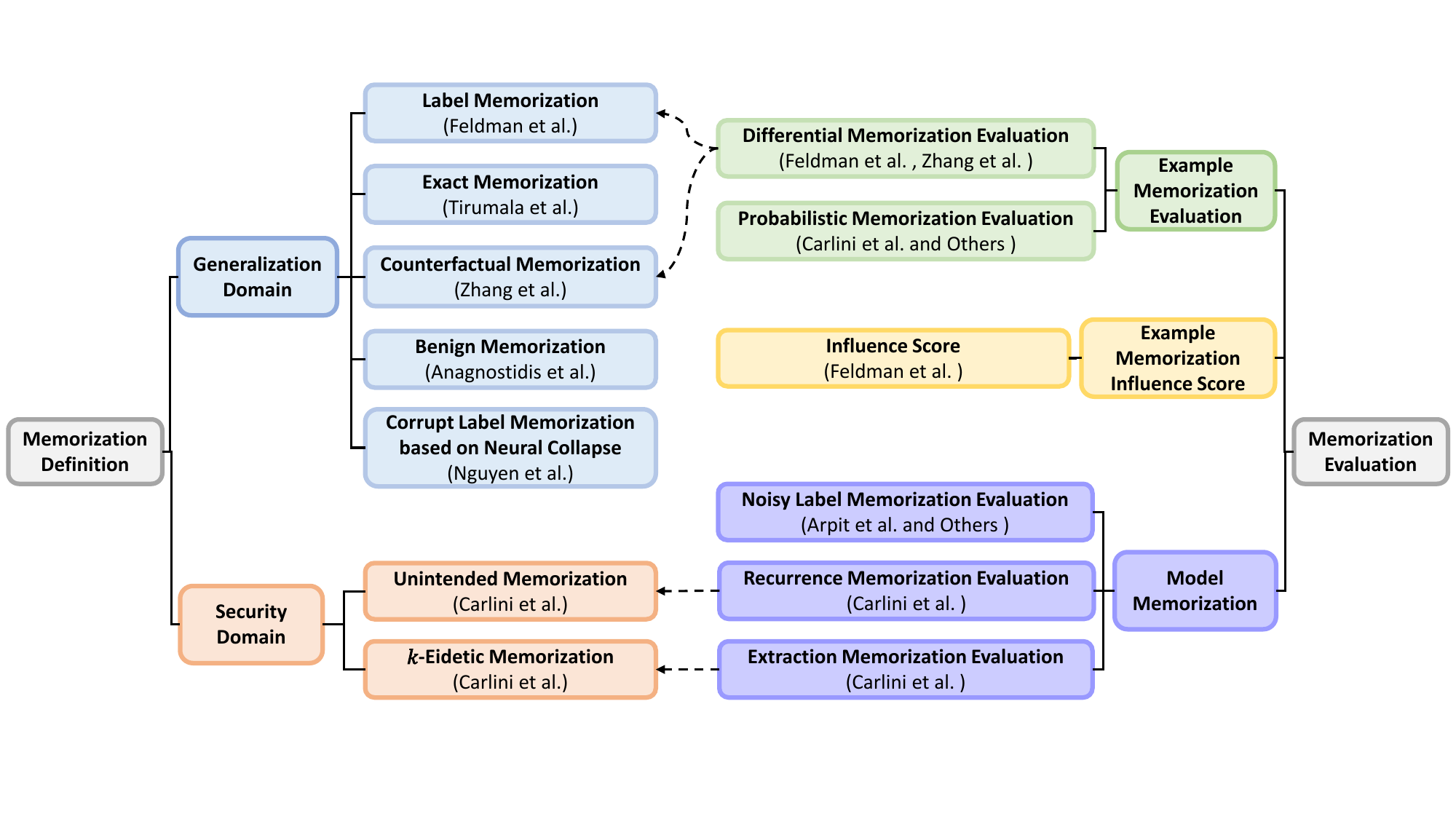}
    \caption{Memorization Definitions and Evaluations.}
    \label{fig:definitionAndMeasurements}
\end{figure*}

\begin{table*}[!htbp]
\centering
\scriptsize
\caption{Main Memorization Definitions}
\label{table:memorizationDifferences}
\begin{tblr}{
  row{1} = {c},
  cells = {c,m},
  colspec = {X[0.5] X[0.7] X[0.7] X[1.5] X[2]},
  cell{2}{1} = {r=5}{c},
  cell{7}{1} = {r=2}{c},
  hline{1-2,7,9} = {-}{1pt},
  hline{3-6,8} = {2-5}{},
}
\textbf{Domain} & \textbf{Name}                                         & \textbf{Reference}       & \textbf{Research Question}                                                              & \textbf{Description}                                                                                                                             \\
Generalization  & Label Memorization                                    & Feldman et al. 2020~\cite{feldman_DoesLearningRequire_2020,feldman_WhatNeuralNetworks_2020}      & Studying memorization effect of long-tailed examples.                                   & This definition provides a universal understanding of
  memorization and distinguishes memorization learning and pattern learning
  effectively. \\
                & Exact Memorization                                    & Tirumala et al. 2022~\cite{tirumala_MemorizationOverfittingAnalyzing_2022}     & Studying underlying training and memorization dynamics of very
  large language models. & The exact memorization actually represents accuracy that
  cannot identify memorization learning in the language model.                          \\
                & Counterfactual Memorization                           & Zhang
  et al. 2021~\cite{zhang_CounterfactualMemorizationNeural_2021}      & Studying counterfactual memorization in language models.                                & {This concept extends label memorization to unsupervised tasks.}                                                                          \\
                & Benign Memorization                                   & Anagnostidis et al. 2023~\cite{anagnostidis_CuriousCaseBenign_2022} & Studying learned features with data augmentation.                                       & Benign memorization describes DNNs can learn useful features
  on the randomly labeled dataset with data augmentation technology.                \\
                & Corrupt Label Memorization
  based on Neural Collapse & Nguyen et al. 2023~\cite{nguyen_MemorizationDilationModelingNeural_2023}       & Studying how corrupt label data impacts neural collapse.                                & The definition attempts to explain the influence of corrupt
  label data in neural collapse.                                                     \\
Security and Privacy        & Unintended Memorization                               & Carlini et al. 2019~\cite{carlini_SecretSharerEvaluating_2019}      & Studying unintended memorization in training.                                           & Unintended memorization definites the features that are irrelevant to the main task but memorized by DNNs.                                        \\
                & $k$-Eidetic Memorization                            & Carlini et al. 2021~\cite{carlini_ExtractingTrainingData_2021}      & Studying privacy leakage in language models.                                            & This memorization definition helps analyze the possibly
    memorized data based on repetition times.                                       
\end{tblr}
\end{table*}

\section{Memorization Definition}~\label{sec:definition}
Memorization is a vague and abstract concept, and difficult to obverse during the training of neural networks. Thus, previous studies did not provide a clear and uniform definition. Based on relevant research, we find that the motivations for studying the memorization phenomenon are its impact on generalization and the concerns about privacy and security risks. In this section, we outline existing definitions of memorization within the contexts of the generalization domain and security domain as 
Table~\ref{table:memorizationDifferences} and Figure~\ref{fig:definitionAndMeasurements}.

% In this section, we provide memorization evaluation at the example level and model level with corresponding distinct evaluation methods. In addition, researchers have proposed specific memorization definitions tailored to various problem domains, and we also expound on these proposed definitions.
%taxonomy

\subsection{Memorization Definitions in Generalization Domain}
\subsubsection{Label Memorization} 
Intuitively, there would exist an obvious disparity when evaluating a data point on a model between the model memorizing the data point and not. Feldman~\cite{feldman_DoesLearningRequire_2020} introduces the label memorization concept to describe the disparity in supervised learning tasks. Label memorization differentially defines what memorizing a label of a point in the dataset means. 

\begin{definition}~\label{def:labelMem}
\textbf{Label Memorization for Supervised Tasks.} Given a training algorithm $A$ that maps a training dataset $D$ to a trained model $f$, the amount of memorization by $A$ on example $(x_i,y_i) \in D$ is defined as
\begin{IEEEeqnarray}{rCl}~\label{eq:mem_score_super}
\resizebox{0.49\textwidth}{!}{$
    \text{mem}(A,D,(x_i,y_i)) := \Pr\limits_{f\leftarrow A(D)}[f(x_i)=y_i] -\Pr\limits_{f^{\prime}\leftarrow A(D^{\backslash i})}[f^{\prime}(x_i)=y_i], \IEEEeqnarraynumspace
$}
\end{IEEEeqnarray}
where $D^{\backslash i}$ denotes the dataset $D$ with $(x_i,y_i)$ removed.
\end{definition}

This definition provides a universal understanding of memorization and distinguishes generalized examples effectively. The definition actually approaches the nature of memorization that memorized examples cannot rely on generalized features. 

\subsubsection{Exact Memorization}
Exact memorization proposed by Tirumala et al.~\cite{tirumala_MemorizationOverfittingAnalyzing_2022}, is used to perform a large-scale study of the dynamics of memorization over training. Additionally, this definition is only applied to the language models.  

\begin{definition}~\label{def:exactMem}
\textbf{Exact Memorization.} Let $V$ denote the vocabulary size. Let $C$ denote a set of contexts, which can be thought of as a list of tuples $(s, y)$ where $s$ is an input context (incomplete block of text) and $y$ is the index of the ground truth token in the vocabulary that completes the block of text. Let $S$ denote the set of input contexts, and let $f:S \rightarrow \mathbb{R}^V$ denote a language model. A context $c =(s, y) \in C$ is memorized if $\arg\max(f(s)) = y$. For a given set of contexts $C$ (i.e., a given training dataset), the proportion of memorized contexts can be represented as:
\begin{IEEEeqnarray}{rCL}
    \text{mem(f)} = \frac{\sum_{(s,y)\in C} 1\{\arg\max(f(s))=y\}}{|C|}.
\end{IEEEeqnarray}
\end{definition}

Based on the formula, we know that the exact memorization actually represents accuracy since it just measures the average number that predicted token matches the ground truth token in the contexts. Thus, this definition is not related to the memorization phenomenon and cannot describe the memorization. 

\subsubsection{Counterfactual Memorization}
Counterfactual memorization is extended from label memorization to unsupervised tasks. Zhang et al.~\cite{zhang_CounterfactualMemorizationNeural_2021} introduce the definition, applying it to quantify the episodic memorization in language models. 
\begin{definition}~\label{def:counterfactualMem} \textbf{Counterfactual Memorization.} Given a training algorithm $A$ that maps a training dataset $D$ to a trained model $f$, and a measure $M$ which measures the performance of $x_i$ on $f$, the amount of memorization by $A$ on example $x_i \in D$ measured with $M$ is defined as
\begin{IEEEeqnarray}{rCL}~\label{eq:mem_score_unsuper}
\resizebox{0.45\textwidth}{!}{$
    \text{mem}(A,D,M,x_i):= \mathop{\mathbb{E}}\limits_{f\leftarrow A(D)}[M(f,x_i)] - \mathop{\mathbb{E}}\limits_{f^{\prime}\leftarrow A(D^{\backslash i})}[M(f^{\prime},x_i)].\IEEEeqnarraynumspace
$}
\end{IEEEeqnarray}

where $M$ can be per-token accuracy that $f$ predicts the next token based on the preceding tokens, then measures the 0-1 loss.
\end{definition}

Basically, counterfactual memorization is a universal version of label memorization and this differential memorization definition can empirically evaluate memorization in various tasks. 

\subsubsection{Benign Memorization}
Benign memorization describes the phenomenon that neural networks can learn useful features on the randomly labeled dataset with data augmentation technology~\cite{anagnostidis_CuriousCaseBenign_2022}. This work regards the general neural network structure as an encoder-projector pair and trains the pair on an augmented noisy dataset. If the accuracy of $k$NN probing at the embedding vectors of the encoder increases over probing at initialization, this is benign memorization.

\begin{definition}~\label{def:benignMem} \textbf{Benign Memorization.} Here are two datasets, $D:={(x_i,y_i)}^n_{i=1}$ denotes the original clean dataset and $\tilde{D} = {(x_i,\tilde{y}_i)}^n_{i=1}$ its randomly labeled version. We call an encoder-projector pair $(h_{\phi*}, g_{\psi*})$ a memorization of $\tilde{D}$, if $f_*$ perfectly fits $\tilde{D}$. Moreover, we call $(h_{\phi*}, g_{\psi*})$ a malign memorization if additionally, probing of $h_{\phi*}$ on $D$ does not improve over probing at initialization. On the contrary, we call $(h_{\phi*}, g_{\psi*})$ a benign memorization of $\tilde{D}$ if probing of $h_{\phi*}$ on $D$ outperforms probing at initialization.
\end{definition}

This definition focuses on the generalization performance when training on randomly labeled datasets. Benign memorization occurs if the encoder learns generalized features. Therefore, this memorization definition is auxiliary to explain noisy label learning rather than a general memorization definition.

\subsubsection{Corrupt Label Memorization based on Neural Collapse}
Empirical evidence indicates that the memorization of noisy data points may lead to degradation (dilation) of the neural collapse. Nguyen et al.~\cite{nguyen_MemorizationDilationModelingNeural_2023} purpose memorization-dilation model and define memorization based on neural collapse under corrupt label training data.

\begin{definition}~\label{def:neuralCollapseMem} \textbf{Corrupt Label Memorization based on Neural Collapse.}  For a given and labeled dataset $D$ with label noise $\eta$ and $K$ categories, if $f$ is a feature extractor, denoting the feature representations $f(x_i^k)$ of the example $x_i^k$ by $h_i^k$. Under neural collapse, any $h_i^k$ will collapse to a single feature representation $h^k$. We denote the set of corrupted instances of class $k$ by $[\tilde{I}^k]$. Memorization can be defined as 

\begin{IEEEeqnarray}{c}
    \text{mem}:=\sum\limits_{k=1}^K\sum\limits_{i\in [\tilde{I}^k]}||h_i^k - h_*^k||
\end{IEEEeqnarray}

where $h_*^k$ denotes the mean of (unseen) test instances belonging to class $k$.
\end{definition}

The DNN under neural collapse intends to map examples with the same ground truth label to a single representation due to the similarity in input features. Therefore, instances of the same ground truth but with randomly corrupted labels lack predictable characteristics, making it challenging for the network to distinguish and separate them in a manner that can be generalized effectively. Thus, when the network is still able to successfully separate such instances, it indicates that the network has memorized the feature representations of the corrupted instances present in the training set. This definition expresses the memorization of noisy examples but only applies to the problem domain of neural collapse. The scope of the definition is limited.

\subsection{Memorization Definitions in Security Domain}

\subsubsection{Unintended Memorization}
Unintended memorization mainly serves to privacy concerns. The concept was first proposed by Carlini et al.~\cite{carlini_SecretSharerEvaluating_2019} when they found that LLMs may memorize some sensitive information like social-security numbers unintentionally. Generally, such memorization is unnecessary for achieving generalization and they give a simple unintended memorization definition.

\begin{definition}~\label{def:unintendedMem} \textbf{Unintended Memorization.}  Unintended memorization occurs when trained neural networks may expose the presence of out-of-distribution training data and the training data is irrelevant to the learning task and definitely does not contribute to improving model accuracy.
\end{definition}

Compared to the differential memorization definitions, the unintended definition focuses specifically on the memorization of out-of-distribution and sensitive data. These data can also be considered as secrets, as they should not be revealed or disclosed by the trained neural networks.

\subsubsection{$k$-Eidetic Memorization}
Carlini et al.~\cite{carlini_ExtractingTrainingData_2021} introduces the concept of $k$-Eidetic Memorization for language tasks. The parameter $k$ represents the count of distinct training examples that contain a specific string.

\begin{definition}~\label{def:kEideticMem} \textbf{$k$-Eidetic Memorization.}  A string $s$ is $k$-eidetic memorized (for $k \geq 1$) by an LM $f_\theta$ if $s$ appears in at most $k$ examples in the training data $X: |{x \in X : s \subseteq x}| \leq k$ and the $s$ is extractable from the LM $f_\theta$ with a prefix $c$ which satisfies 
\begin{IEEEeqnarray}{c}
    s \leftarrow \arg\max\limits_{s^{\prime}:|s^{\prime}|=N} f_{\theta}(s^{\prime}|c)
\end{IEEEeqnarray}
where $f_{\theta}(s^{\prime}|c)$ is the likelihood of an entire sequence $s^{\prime}$ with length $N$.
\end{definition}

This memorization definition helps figure out the possibly memorized strings based on repetition times in LM. If $k$ is large, the memorized string may be common knowledge like the zip code of a particular city. But when $k$ is very small, the memorized string could be harmful like accidentally exposing a personal phone number. The $k$-Eidetic Memorization is also concerned with privacy but utilizes the repetition times as a parameter to identify common knowledge memorization and harmful unintended memorization in language tasks.

\section{Memorization Evaluations}~\label{sec:eval}
Memorization evaluation basically follows different memorization levels to identify the existence of memorization and its influence. Figure~\ref{fig:definitionAndMeasurements} demonstrates the methods of memorization evaluation and associated memorization definitions. 

\subsection{Example Memorization Evaluation}
Example memorization focuses on individual example memorization, which means checking if any example has been memorized by neural networks.

\subsubsection{Differential Memorization Evaluation} 
When example memorization happens, the model's outputs on the memorized data between the model training with it and without it will produce a large gap. This is 'leave-one-out' memorization or differential memorization that we present in Defintion~\ref{def:labelMem} and Definition~\ref{def:counterfactualMem}. The measurement can be called the memorization score and the formula is the same as the definitions. 

The memorization score for supervised tasks is 
\begin{IEEEeqnarray*}{rCl}
    \resizebox{0.45\textwidth}{!}{
    $
        \text{mem}(A,D,(x_i,y_i)) := \Pr\limits_{f\leftarrow A(D)}[f(x_i)=y_i] - \Pr\limits_{f^{\prime}\leftarrow A(D^{\backslash i})}[f^{\prime}(x_i)=y_i]
    $
    }   
\end{IEEEeqnarray*}

and the memorization score for unsupervised tasks is 
\begin{IEEEeqnarray*}{rCl}
    \resizebox{0.45\textwidth}{!}{
    $
        \text{mem}(A,D,M,x_i):= \mathop{\mathbb{E}}\limits_{f\leftarrow A(D)}[M(f,x_i)] - \mathop{\mathbb{E}}\limits_{f^{\prime}\leftarrow A(D^{\backslash i})}[M(f^{\prime},x_i)].
    $
    }   
\end{IEEEeqnarray*}

The memorization score quantifies the performance gap on a single example when the example is included and excluded from the training dataset. A notably large performance gap indicates that other examples cannot provide useful features for the data example and the model has to memorize it. Additionally, this measurement may require more computation resources. Jiang et al.~\cite{jiang_CharacterizingStructuralRegularities_2021} provide a simplified method to use multiple large subsets to replace the held-out dataset which saves the evaluation cost.

\subsubsection{Probabilistic Memorization Evaluation}~\label{evl:prob} 
Probabilistic memorization depends on the differences in model outputs of memorized and generalized examples. There may exist multiple techniques to capture the differences but the most relevant method is the membership inference attack.

This kind of attack aims to determine whether a data point belongs to the training dataset. The success of the attack cannot rely on the generalized feature of examples because these features are common for the entire data distribution. Therefore, the membership inference attack focuses on the particular or unique features that models memorize. In other words, data points that the model has memorized during training are more likely to be correctly identified as belonging to the training dataset in membership inference attacks. Though there is no obvious quantitative research to prove such a relationship and no formal definition, some works~\cite{carlini_PrivacyOnionEffect_2022,li_PrivacyEffectData_2023} tacitly approve the relationship and adopt membership inference attack to measure memorization. Thus, it is possible to use membership inference attacks to evaluate model memorization indirectly. It is noted that some membership inference methods have high false positive rates which cannot exactly measure memorization~\cite{rezaei_DifficultyMembershipInference_2021,carlini_MembershipInferenceAttacks_2022}.

The typical work is \textit{Likelihood Ratio Attack (LiRA)}~\cite{carlini_MembershipInferenceAttacks_2022}. The core idea behind LiRA is similar to \textbf{Definition}~\ref{def:labelMem}, which involves evaluating membership inference risks by leveraging likelihood ratios. LiRA aims to assess whether a given data point is a member of the training dataset by computing the likelihood ratio based on the model's predictions of the data point when the training dataset includes and excludes it. The original formula can be demonstrated as 
$$
\Lambda(f,(x_i, y_i)) = \frac{p(f|\mathbb{Q}_{in}(x_i, y_i))}{p(f|\mathbb{Q}_{out}(x_i, y_i))}
$$ 
where $\mathbb{Q}_{in}(x_i, y_i)$ and $\mathbb{Q}_{out}(x_i, y_i)$ represents the distribution of models trained on the training dataset with and without the data point $(x_i,y_i)$ and $p$ is the probability density function over $f$ under the distribution of model parameters $\mathbb{Q}$.
The similarity in the core idea highlights the connection between membership inference risks and memorization evaluation.

Moreover, there may exist other techniques based on probability that can be used to estimate memorization. Nevertheless, relevant techniques need careful validation and confirmation that they can really reflect the memorization effect.

\subsection{Example Memorization Influence Score}
The influence score represents how an individual memorized example impacts model generalization performance. 

Now, we know some methods to evaluate example memorization but we are also curious about how memorized examples influence model generalization. To measure the impact of an individual memorized example on generalization, the influence score based on the memorization score has been proposed~\cite{feldman_WhatNeuralNetworks_2020}. Generally, the influence score of a training example $(x_i,y_i)$ against a test example $(x_j^{\prime},y_j^{\prime})$ under a supervised task can be defined as 
\begin{IEEEeqnarray}{rCl}
    \resizebox{0.45\textwidth}{!}{$
        \text{infl}(A,D,(x_i,y_i),(x_j^{\prime},y_j^{\prime})) := \Pr\limits_{f\leftarrow A(D)}[f(x_j^{\prime})=y_j^{\prime}] - \Pr\limits_{f^{\prime}\leftarrow A(D^{\backslash i})}[f^{\prime}(x_j^{\prime})=y_j^{\prime}].
    $}
\end{IEEEeqnarray}

 Similarly, the influence score for unsupervised tasks~\cite{zhang_CounterfactualMemorizationNeural_2021} can be defined as 
 \begin{IEEEeqnarray}{rCl}
    \resizebox{0.45\textwidth}{!}{$
        \text{infl}(A,D,M,x_i,x_j^{\prime}) := \mathop{\mathbb{E}}\limits_{f\leftarrow A(D)}[M(f,x_j^{\prime})] - \mathop{\mathbb{E}}\limits_{f^{\prime}\leftarrow A(D^{\backslash i})}[M(f^{\prime},x_j^{\prime})].
    $}
\end{IEEEeqnarray}
In the corresponding work~\cite{feldman_WhatNeuralNetworks_2020,zhang_CounterfactualMemorizationNeural_2021}, they find a direct positive correlation between memorization scores and influence scores, and these examples are almost atypical. This observation proves rare and memorized examples provide particular features for their subpopulation generalization. Moreover, it's worth noting that not all memorized examples contribute high influence scores because some memorized examples can be regarded as noisy examples and some are so rare that even no test examples belong to the corresponding subpopulation.

\begin{figure*}[t]
    \centering
    \includegraphics[width=0.8\textwidth]{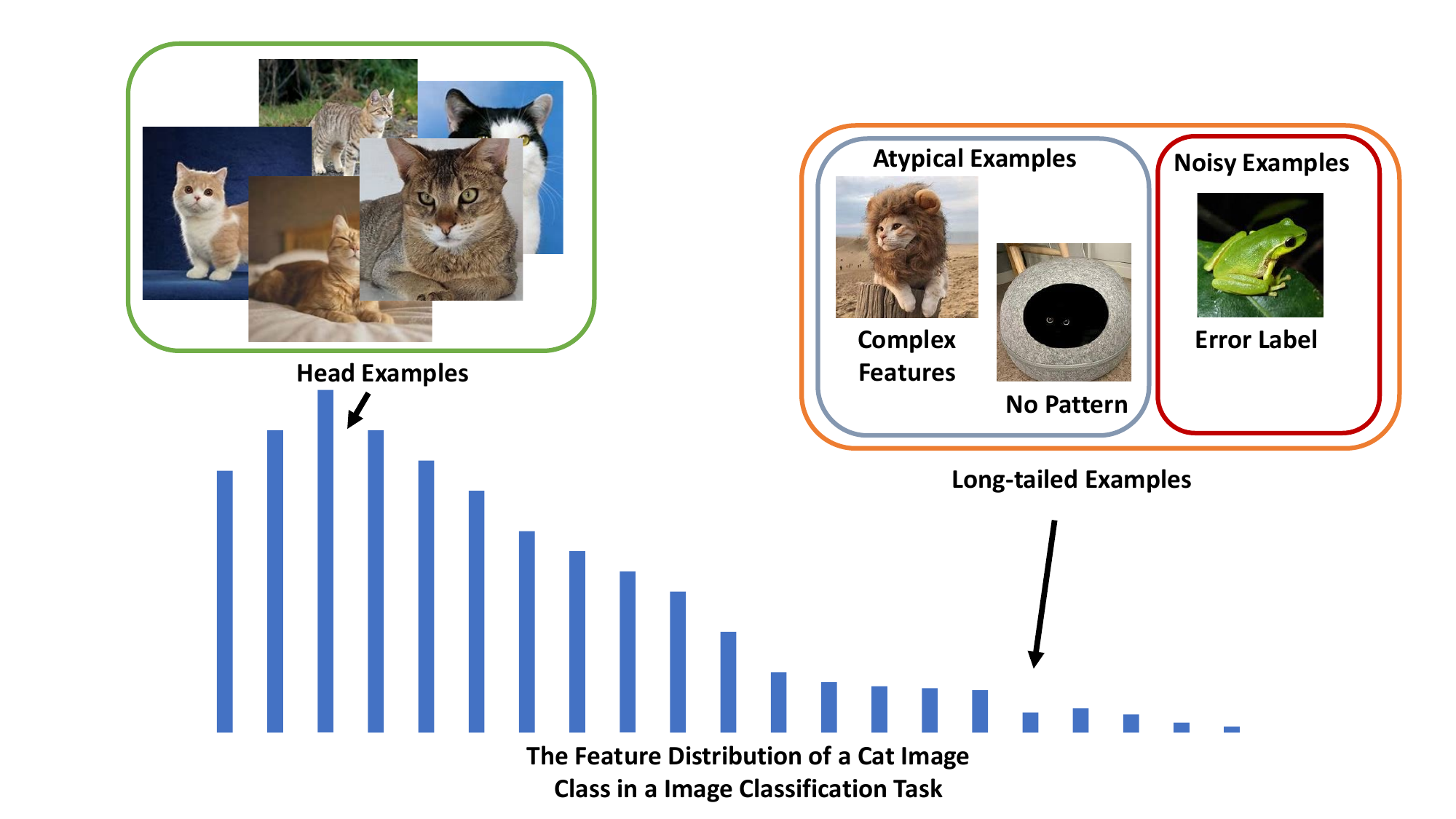}
    \caption{Demonstration of the Long-tailed Examples.}
    \label{fig:longtailedExamples}
\end{figure*}

\subsection{Model Memorization Evaluation}
Model memorization cares about the role of neural networks, concerning how much memorization exists in models and the memorization capacity and ability of models. 

\subsubsection{Noisy Label Memorization Evaluation}
Noisy label memorization evaluation actually is not used to measure model memorization but is a valuable method to build memorization baselines compared to other properties of the model. Depending on the fact that noisy label examples have no shared class-level features and patterns, the model has to memorize all of these noisy label examples. Thus, many works utilize noisy label examples as known memorization. Arpit et al.~\cite{arpit_CloserLookMemorization_2017} mix the noisy label examples with normal examples to study the learning dynamics during training. They use the ratio of noisy label examples in the training dataset to represent memorization. Another work~\cite{dong_ExploringMemorizationAdversarial_2022} is studying the memorization effect in adversarial training, utilizing the randomly labeled adversarial examples. Additionally, Maini et al.~\cite{maini_CanNeuralNetwork_2023} attempt to employ noisy label examples to localize the memorization in the neural network. Hence, noisy label memorization evaluation is a common method to investigate the relationships between memorization and other factors. 

\subsubsection{Recurrence Memorization Evaluation} 
Recurrence memorization evaluation refers to the probability that neural networks can generate or extract specific marked examples put in the training dataset to measure the memorization tendency and ability of the model. Obviously, the selection of marked examples mainly affects the memorization evaluation. 

Carlini et al.~\cite{carlini_SecretSharerEvaluating_2019} employ random sequences to evaluate unintended memorization (Definition~\ref{def:unintendedMem}) in language models depending on this evaluation method. Specifically, they build the canary sequences which consist of two parts. The first part is like "the random number is" and the second part is just random numbers. Consequently, they create a metric called exposure index based on the log-perplexity, 
$$
    \mathbf{Px}_{f}(x_1,\cdots,x_n) = \sum_{i=1}^N (- \log_2 \mathbf{Pr}(x_i|f(x_1,\cdots,x_{i-1}))),
$$
where $f$ is the language model, and $x_1,\cdots,x_n$ represents the input sequence. The perplexity measures how “surprised” the model is to see a given value. A higher perplexity indicates the model is “more surprised” by the sequence. Therefore, the exposure index measures the likelihood of data sequences. The evaluation follows confirming the canary sequence inserted into the training dataset, training, and then applying the exposure index to gain the probability of the canary sequence reproduction. The exposure index of the canary sequence may represent the model memorization ability. 

Additionally, employing other types of examples like atypical examples instead of random examples may disclose other properties of model memorization. This requires further studies.

\subsubsection{Extraction Memorization Evaluation} 
Extraction memorization evaluation applies the extraction or inversion approaches to empirically evaluate the model memorization by producing all extractable examples and identifying those in the training dataset. This method attempts to provide a lower bound of model memorization. However, it is noted that not all extracted examples are identical to corresponding training examples because memorization works on the feature scale. Some extractable examples can be regarded as the representatives of generalized examples. It requires good metrics to ensure extractable examples are really memorized.

An effective work~\cite{carlini_ExtractingTrainingData_2021} applying this method attempts to extract training data from large language models and find examples with small $k$ in $k$-Eidetic Memorization (Definition~\ref{def:kEideticMem}). They generate a lot of text with GPT-2 and pick text with the highest memorization probability, validating the memorization on picked text manually via Internet search.

This evaluation method may require more resources but perhaps additionally assist researchers in understanding what models memorize.

\section{Memorization in DNN Training}~\label{sec:memorizationBehaviors}
A primary motivation behind research on memorization is to explore its impact and what role it plays in DNN training. In this section, we will provide a comprehensive review of memorization research in the DNN training framework. Table~\ref{table:DNNinTraining} demonstrates the main relevant works in this area.

\begin{table*}[!htbp]
\centering
\scriptsize
\caption{Main Works about Memorization in DNN Training}
\label{table:DNNinTraining}
%\fontsize{4}{4}\selectfont
%\resizebox{1\linewidth}{!}{
\begin{tblr}{
  cells = {c,m},
  colspec = {X[1.2] X[1] X[1.5] X[1.5] X[3]},
  cell{3}{1} = {r=5}{},
  cell{8}{1} = {r=2}{},
  cell{10}{1} = {r=3}{},
  cell{14}{1} = {r=2}{},
  hline{1-3,8,10,13-14,16} = {-}{1pt},
  hline{-} = {-}{},
}
\textbf{Main Topic}                       & \textbf{Reference}                          & \textbf{Background Task}                       & \textbf{Main Eval. Method}                                  & \textbf{Main Findings}     \\
{Memorization Mechanism}                     & Zhang et al., 2017~\cite{zhang_UnderstandingDeepLearning_2017}                  & {Supervised \\ Classification Task}        & {Noisy Label Memorization Evaluation}                   & DNNs can memorize randomly labeled datasets that traditional approaches
  fail to explain generalization.                                                  \\
{Memorization about Data}                  & Feldmen et al., 2020~\cite{feldman_DoesLearningRequire_2020,feldman_WhatNeuralNetworks_2020}                & {Supervised \\ Classification Task}        & {Differential Memorization Evaluation}                  & Propose the long tail theory that memorization of long-tailed examples is
  crucial for achieving close-to-optimal generalization error.                    \\
                                           & Hacohen et al., 2020~\cite{hacohen_LetAgreeAgree_2020}               & Various Tasks                         & /                                                     & Different neural networks memorize data in different orders.                                                                                               \\
                                           & Zhang et al., 2021~\cite{zhang_CounterfactualMemorizationNeural_2021} & {Unsupervised Language \\ Generative Task} & {Differential emorization Evaluation}                  & High memorization examples are generally unconventional texts.                                                                                             \\
                                           & Lee et al., 2022~\cite{lee_DeduplicatingTrainingData_2022}                    & {Unsupervised Language \\ Generative Task} & {Probabilistic Memorization Evaluation}                 & Deduplicated datasets make less memorization.                                                                                                              \\
                                           & Carlini et al., 2023~\cite{carlini_QuantifyingMemorizationNeural_2023} & {Unsupervised Language \\ Generative Task} & {Extraction Memorization Evaluation}                    & Repeated examples have a high probability of being extracted.                                                                                              \\
{Memorzation about Training Stage}         & Arpit et al., 2018~\cite{arpit_CloserLookMemorization_2017}                  & {Supervised \\ Classification Task}        & {Noisy Label Memorization Evaluation}                   & Learning simple patterns is prior to remembering noise data in the early
  training stage.                                                                 \\
                                           & Maennel et al., 2020~\cite{maennel_WhatNeuralNetworks_2020}                & {Supervised \\ Classification Task}        & {Noisy Label Memorization Evaluation}                   & An alignment between the
  principal components of network parameters and data takes place when training
  with random labels in the early training stage. \\
{Memorization about Architecture }         & Stephenson et al., 2021~\cite{stephenson_GeometryGeneralizationMemorization_2021}             & {Supervised \\ Classification Task}        & {Mean Field \\ Theoretic Geometric Analysis} & Memorization predominately
  occurs in the deeper layers.                                                                                                  \\
                                           & Maini et al., 2023~\cite{maini_CanNeuralNetwork_2023}                  & {Supervised \\ Classification Task}        & {Noisy Label Memorization Evaluation}                   & Memorization exists in a small
  set of neurons in various layers of the model.                                                                            \\
                                            &Geva et al., 2021~\cite{geva_TransformerFeedForwardLayers_2021}	& {Unsupervised Language Generative Task}	& /	& Feed-forward layers in Transformer are key-value memories. \\
{Memorization about Overfitting }            & Tirumala et al., 2022~\cite{tirumala_MemorizationOverfittingAnalyzing_2022}               & {Unsupervised Language \\ Generative Task} & Exact Memorization                                    & Larger models can memorize a
  larger portion of the data before over-fitting                                                                              \\
{Memorization about DA and Regularization} & Anagnostidis et al., 2023~\cite{anagnostidis_CuriousCaseBenign_2022}           & {Supervised \\ Classification Task}        & $k$NN Probe                                             & Even randomly labeled datasets with DA could lead to highly useful
  features.~                                                                            \\
                                           & Li et al., 2023~\cite{li_PrivacyEffectData_2023}                     & {Supervised \\ Classification Task}        & {Probabilistic Memorization Evaluation}                 & Trivial data augmentation
  technologies can mitigate memorization.                                                                                        
\end{tblr}
%}
\end{table*}

\subsection{Exploring Memorization Mechanism}
In some early studies~\cite{krueger*_DeepNetsDon_2017}, researchers believed that the memorization effect was not necessary for learning. Generally, the generalization of DNNs means networks can learn and recognize common patterns hidden within the input data. These common patterns consist of shared features among similar data examples. When DNNs learn these common patterns, they exhibit the ability to generalize, thereby demonstrating their capacity to perform well on new, unseen data beyond the training dataset. In contrast, memorization means networks memorize input examples rather than patterns which results in overfitting. However, as Zhang et al.~\cite{zhang_UnderstandingDeepLearning_2017} find that DNNs can easily fit the random labeled dataset, the traditional statistical learning theory like VC dimension~\cite{vapnik1998adaptive}, Rademacher complexity~\cite{bartlett2001rademacher}, and uniform stability~\cite{mukherjee2002statistical,bousquet2002stability,poggio2004general} cannot explain the generalization of DNNs. It is well known that DNNs cannot correctly classify the randomly labeled dataset based on the common patterns of the data distribution, thus DNNs have to memorize the entire dataset. However, the memorization mechanism in the DNNs remains unclear and vague. Therefore, two important and interesting questions arise: why DNNs memorize data in the standard training process, and how the memorization mechanism operates. This attracts the machine learning community to explore the memorization effect.

 % However, as DNN can successfully demonstrate good generalization on various tasks, Zhang et al.~\cite{zhang_UnderstandingDeepLearning_2017} find that DNNs can easily fit a random labeling of the training data and traditional explanations like properties of the model family, or the regularization techniques used during training fails to explain why large neural networks generalize well in practice. 

\subsection{Memorization and Data Training}
In studying overfitting~\cite{goodfellow2016deep},  researchers have found that DNNs may memorize data. In exploring the memorization phenomenon, understanding the relationship between data distribution and memorization tendency, orders, as well as how the memorization mechanism affects training performance, is an important step.

The real-world natural data distributions are generally long-tailed~\cite{reed_ParetoZipfOther_2001} and almost all practical datasets are sampled from the real world. Considering this, Feldman et al.~\cite{feldman_DoesLearningRequire_2020,feldman_WhatNeuralNetworks_2020} propose the long tail theory. This theory suggests that long-tailed examples as illustrated in Figure~\ref{fig:longtailedExamples} are prone to be memorized. Moreover, memorizing these long-tailed examples is crucial for achieving close-to-optimal generalization errors in long-tailed data distributions because rare and atypical instances can provide necessary generalization. 
%Additionally, in such distributions, valuable examples from the "long tail" can be statistically indistinguishable from the useless ones, such as outliers and mislabeled instances. 
They further validate the theory by evaluating examples based on the memorization score (Definition~\ref{def:labelMem}). The results illustrate that examples with high memorization scores are more atypical. Thus, compared to typical examples, atypical examples are more likely to be memorized by DNNs. When removing examples with high memorization scores, the generalization errors increase. This theory also has empirical evidence in language tasks~\cite{zheng_EmpiricalStudyMemorization_2022}. Jiang et al.~\cite{jiang_CharacterizingStructuralRegularities_2021} develop the consistency score (C-score) based on memorization score (Definition~\ref{def:labelMem}), which can be applied in larger datasets. The C-score aims to measure the per-instance generalization. Their result demonstrates that a more atypical example has a lower C-score which provides convincing evidence for the long tail theory. Zhang et al.~\cite{zhang_CounterfactualMemorizationNeural_2021} extend the memorization score to unsupervised learning and propose counterfactual memorization (Definition~\ref{def:counterfactualMem}) to evaluate text datasets. They discover that high memorization examples are generally unconventional texts such as all-capital letters, structured formats, and multilingual texts. In contrast, low memorization examples are generally templated documents with many near-duplicate copies in the training data. The tendency also corresponds to the long tail theory~\cite{feldman_WhatNeuralNetworks_2020}. 

Moreover, a recent research~\cite{maini_CanNeuralNetwork_2023} indicates that the DNNs cannot identify noisy examples from atypical examples empirically. When removing memorization-associated neurons, DNNs cannot classify the noisy examples and the generalization performance also reduces on the noise-mixed dataset. Additionally, for the same memorized example, the memorization path is distinct in repeated independent training experiments~\cite{gu_NeuralNetworkMemorization_2019}. This may represent that DNNs can select various particular features to uniquely identify the same example. Furthermore, the learning order of clean examples has observable consistency in similar architectures~\cite{hacohen_LetAgreeAgree_2020}. However, when training DNNs on the same dataset with randomly shuffled labels, they find that different models memorize the data in a different order. This finding also suggests that memorization learning has various possible paths. 

\textbf{Summary.} Atypical and noisy examples as long-tailed examples lack representativity in datasets, thus, DNNs are prone to memorize these long-tailed examples to minimize the training loss. This explains why and what DNNs memorize.

\subsection{Memorization and Repetition}

Intuitively, DNNs tend to memorize duplicated examples. Zhang et al.~\cite{zhang_CounterfactualMemorizationNeural_2021} believe most memorization criteria strongly correlate with the number of example occurrences in the training, and language models will capture common memorization such as familiar phrases, public knowledge, or templated texts. Moreover, deduplicated datasets reduce the memorization frequency and improve generalization~\cite{lee_DeduplicatingTrainingData_2022}. From the perspective of the extraction task, repeated examples have a high probability of being extracted~\cite{carlini_QuantifyingMemorizationNeural_2023}.  

One study~\cite{carlini_ExtractingTrainingData_2021} links repetition times and memorization, proposing $k$-Eidetic Memorization (Definition~\ref{def:kEideticMem}), where $k$ relates to the number of occurrences for one example. They apply this definition to the language model extraction task and investigate GPT-2. For extractable large $k$ examples, they include common knowledge like city names or high-frequency words, and complex text such as the entire text of the MIT public license because the license may occur thousands of times in the training dataset. However, GPT-2 also memorizes some low-frequent examples with small $k$ like contact information and valid URLs.  

Therefore, in practical environments, example repetition is an influence factor of memorization. Nevertheless, the long tail theory~\cite{feldman_DoesLearningRequire_2020,feldman_WhatNeuralNetworks_2020} tells us that the long-tailed examples are prone to be memorized and these long-tailed examples are low-frequent in the distribution. This requires systematically evaluating memorization factors.   

\textbf{Summary.} DNNs tend to prioritize the memorization of repeated data. However, memorization learning is limited by multiple factors. There is currently no unified framework to describe the impact of data on memorization.

\subsection{Memorization and Training Stage}

Researchers have discovered that DNN training has a critical early learning stage~\cite{frankle_EarlyPhaseNeural_2019,achille2018critical}, during which model performance increases rapidly. Then with the truth that DNNs typically minimize loss in the final stage of training~\cite{zhang_UnderstandingDeepLearning_2017,papyan2020prevalence}, it is reasonable to believe that pattern learning and memorization learning dominate different training stages. Therefore, investigating and explaining the memorization dynamic during training stages constitutes a valuable research topic.

Due to the difficulty in separating the memorization learning from the generalization, it is possible to explore how DNN learns by using a training dataset containing a mixture of normal and noisy examples in certain proportions. Arpit et al.~\cite{arpit_CloserLookMemorization_2017} utilize the method and find that DNNs tend to prioritize learning patterns even in noisy datasets, as evidenced by high validation accuracy in the early training stage. Subsequently, DNNs begin to directly memorize noisy examples, leading to a rapid drop in validation accuracy. Maennel et al.~\cite{maennel_WhatNeuralNetworks_2020} obverse an alignment between the principal components of network parameters and data takes place when training with random labels in the early training stage. However, the misalignment scores gradually increase during the later training stage.

Another perspective~\cite{stephenson_GeometryGeneralizationMemorization_2021,gu_NeuralNetworkMemorization_2019} based on analyzing the gradient variation explains the phenomenon. In the early learning phase, the gradients from noisy examples contribute minimally to the total gradient because inconsistent gradient information may counteract each other, and those shared patterns of the same class are consistent, facilitating quick updates and promoting pattern learning. Similarly, applying a new detection method called 'variance of gradients' (VoG)~\cite{agarwal_EstimatingExampleDifficulty_2022}, the examples with lower VoG in the early training stage are more typical compared to the examples in the later training stage. Combining this with the long tail theory~\cite{feldman_DoesLearningRequire_2020}, it may be inferred that pattern learning dominates the early training stage. 

\textbf{Summary.} 
There is sufficient evidence to demonstrate that pattern learning dominates the early training stage, while memorization learning mainly occupies the relatively later training stage. One reasonable explanation is the subpopulation with the same pattern can contribute effective gradients during the early learning stage. Conversely, atypical examples and noisy examples contain conflicting gradient information that cannot be effectively learned during early training.

\subsection{Memorization and Model Architecture}

Different network layers experience diverse learning dynamics. In exploring the functions of DNN layers, a study on transferability~\cite{yosinski_HowTransferableAre_2014} suggests that shallow layers' features appear to be general and applicable to other tasks or datasets. However, deep layers tend to learn task-correlated features. This illustrates that different layers learn distinct features, with shallow layers in DNNs being more prone to learn patterns while deep layers specialize.

Some subsequent studies have contributed to enhancing the reliability of this viewpoint. Using Singular Vector Canonical Correlation Analysis (SVCCA), Raghu et al.~\cite{raghu_SVCCASingularVector_2017} compare layers across time and observe their convergence starting from the shallow layers. Morcos et al.~\cite{morcos_InsightsRepresentationalSimilarity_2018} develop projection-weighted Canonical Correlation Analysis (CCA) based on SVCCA. With multiple networks, they demonstrate that generalized networks converge to more similar representations. Specifically, they note that at shallow layers, all networks converged to equally similar solutions. Intuitively, this indicates that the shallow layers learn common patterns. However, at deep layers, groups of generalizing networks converged to substantially more similar solutions compared to groups of memorizing networks. This indicates that each memorizing network memorizes the training data using different strategies. Ansuini et al.~\cite{ansuini_IntrinsicDimensionData_2019} employ the Intrinsic Dimensionality (ID) of data representations, i.e. the minimal number of parameters needed to describe a representation. In their study on the utility of different layers, they find across layers, the ID initially increases and then progressively decreases in the final layers. Remarkably, they observed that the ID of the last hidden layer could predict classification on the test set.

Clearly, the specialization of deep layers connects to memorization learning. Therefore, a reasonable inference is that the memorization of training examples occurs in the last (or final few) layers of a deep network. Stephenson et al.~\cite{stephenson_GeometryGeneralizationMemorization_2021}, employing replica-based mean field theoretic geometric analysis method, believe memorization mainly occurs in the deeper layers due to decreasing object manifolds’ radius and dimension, whereas previous layers are minimally affected. In the experiment, if rewinding the parameters of the final convolutional layer to an earlier epoch, the generalization of the model can achieve a similar performance to the early-stopped model. Moreover, Anagnostidis et al.~\cite{anagnostidis_CuriousCaseBenign_2022} employ the $k$NN probing to evaluate feature learning of each network layer, utilizing embedding vectors of each training example from each layer and correct labels to build $k$NN models. They find that embedding vectors of test examples produced from shallow layers achieve non-trivial $k$NN accuracy with the randomly labeled training dataset and data augmentation. They term this phenomenon benign memorization (Definition~\ref{def:benignMem}). However, the $k$NN probing accuracies drop significantly in the deep layers, indicating that DNN fits the random labels. This suggests that only the very last layers are used for memorization, while previous layers encode generalized features that remain largely unaffected by the label noise. Furthermore, another work~\cite{maennel_WhatNeuralNetworks_2020} explains the training data will align the principal components of network parameters at the earlier layers when trained with random labels and later layers become specialized. 

However, the latest work conducted by Maini et al.~\cite{maini_CanNeuralNetwork_2023} reveals that memorization of a classification task exists in a small set of neurons in various layers of the model, and the layers that contribute to example memorization are, not the final layers. In their experiment, they use a noisy dataset to train DNNs. Subsequently, they apply technologies known as layer retraining and layer rewinding to eliminate memorization within individual layers. Finally, they validate the memorization effect (i.e. accuracy of noisy examples in the training dataset) on modified models. The unexpected finding is that the memorization effect still persists in the model, which proves various layers contribute to the memorization. 

Furthermore, researchers also investigate the inherent functions of layers regarding memorization. An interesting work~\cite{chatterjee_LearningMemorization_2018} attempts to demonstrate if only the memorization function can provide generalization. The author builds a network of lookup tables and finds deep look-up table network exhibits generalization in the Binary-MNIST task. Moreover, this model reproduces a crucial finding with neural networks: memorization can provide generalization with depth, though it is doubtful that the model can work on more complex tasks. Additionally, Zhang et al.~\cite{zhang_IdentityCrisisMemorization_2020} investigate whether different trained networks tend to demonstrate the constant function (memorization) or the identity function (generalization) and they empirically find that different architectures exhibit strikingly different complex biases. 

As Transformers~\cite{vaswani_AttentionAllYou_2017a} achieve a big success in various tasks, people are also interested in memorization of Transformers and large language models. Sukhbaatar et al.~\cite{sukhbaatar_AugmentingSelfattentionPersistent_2019} augment the self-attention layers with persistent memory vectors and find this plays a similar role as the feed-forward layer. Moreover, Geva et al.~\cite{geva_TransformerFeedForwardLayers_2021} directly point out that feed-forward layers are key-value memories, where each key correlates with textual patterns in the training examples, and each value induces a distribution over the output vocabulary. It should be noted here that this kind of key-value memory combines pattern learning and memorization learning. They show that feed-forward layers act as pattern detectors over the input across all layers and learned patterns are human-interpretable. Additionally, Dai et al.~\cite{dai_KnowledgeNeuronsPretrained_2022} introduce the concept of knowledge neurons that express factual knowledge and propose a knowledge attribution method to identify the neurons via a fill-in-the-blank cloze task in BERT. They find that the activation of such knowledge neurons is positively correlated to the expression of their corresponding facts.

\textbf{Summary.} Functions of different layers in neural networks vary significantly. While many works prove deep layers specialize, memorization location still requires more exploration. For Transformers, researchers have found that feed-forward layers are key-value memories but the memory is not only the result of memorization learning. Among these key-value memories, we cannot ensure they are all task-correlated. Some irrelevant and unexpected details could also be stored in them. Further research is needed to investigate the memorization mechanism associated with the model architecture.

% \textbf{Summary.} According to existing research, memorization seems like an instinctive feature of DNNs. In the training, shared features are prior for the network to learn and these features or patterns can help typical and common examples archive the minimum training loss. But for those long-tail examples or outliers and corrupt examples, just shared features or easy patterns cannot help reduce training loss, thus networks will attempt to learn more features until memorize these examples in various layers. For long-tail examples, memorization is necessary to improve generalization~\cite{feldman_DoesLearningRequire_2020,feldman_WhatNeuralNetworks_2020}. Otherwise, memorized outliers and noisy examples may confuse DNNs and hurt generalization~\cite{arpit_CloserLookMemorization_2017,anagnostidis_CuriousCaseBenign_2023}. 

\subsection{Memorization and Overfitting}

Overfitting is a common phenomenon in deep learning which represents that a model learns the training data so well that it captures not only the underlying patterns but also the particular features in the data. This causes the model to fail in generalizing effectively to new, unseen data. Thus, early research commonly held the opinion that overfitting was responsible for memorization. However, contemporary studies~\cite{carlini_SecretSharerEvaluating_2019,maini_CanNeuralNetwork_2023,tirumala_MemorizationOverfittingAnalyzing_2022} provide evidence supporting the persistence of memorization throughout the training process. Memorization does not necessarily lead to overfitting.

Based on the long tail theory~\cite{feldman_DoesLearningRequire_2020,feldman_WhatNeuralNetworks_2020}, we understand that memorizing atypical examples contributes to generalization. In contrast, overfitting will enlarge the generalization error while training loss decreases. Some recent works suggest that even for DNNs without a significant train-test gap, memorization still exists~\cite{carlini_ExtractingTrainingData_2021,tirumala_MemorizationOverfittingAnalyzing_2022}. Additionally, the privacy risks (Evaluation~\ref{evl:prob}) also imply the underlying memorization. It is known that overfitting is not necessary for successful membership inference attacks~\cite{yeom_PrivacyRiskMachine_2018,long_UnderstandingMembershipInferences_2018}. Furthermore, when a neural network is trained in a training dataset mixed with clean examples and less noisy examples, both the accuracy of noisy examples and that of clean examples exhibit concurrent improvement~\cite{maini_CanNeuralNetwork_2023}. Overfitting is a phenomenon of training observed in the later stages of training. For individual training examples, DNNs may memorize them while learning patterns in the early training stage. Thus, memorization does not necessarily require overfitting.

%Benign Overfitting 
Another interesting phenomenon is benign overfitting. The phenomenon means that even after overlearning training data, DNNs still can generalize well~\cite{bartlett_BenignOverfittingLinear_2020,li_UnderstandingBenignOverfitting_2021}. This theory believes overparameterized DNNs can generalize to the majority of the data distribution using simple paths, and memorize mislabeled and irregular data using complex paths. These components do not interfere, making such overfitting benign. One explanation~\cite{cao_BenignOverfittingTwolayer_2022} believes that overfitting becomes benign when the signal-to-noise ratio satisfies a certain condition. In simple terms, benign overfitting requires sufficient signals in the dataset. Thus, benign overfitting may involve less memorization, yet there is insufficient evidence to illustrate their relationship.

\textbf{Summary.} Overfitting as a training phenomenon does not have a strong relationship with memorization. In the context of overfitting, memorization is necessary but not sufficient.

\subsection{Memorization and Data Augmentation, Regularization}
Data augmentation and regularization are widespread techniques used in training neural networks. Therefore, it is necessary to study the impact of these practices on memorization.

\paragraph{General Data Augmentation}
Data Augmentation is a pivotal strategy used to expand the original training dataset by introducing a variety of artificially generated examples. Generally, the primary objective of this technique is to enrich example representations based on corresponding semantic features. Multiple representations can help DNNs perform well on unseen examples, thereby improving the generalization. Presently, various technologies exist to implement data augmentation, but here we focus on trivial data augmentation, which means fundamental transformations applied to the original training dataset. Trivial data augmentation depends on specific data formats. For instance, in image processing, these transformations could include rotations, flips, zooms, and color variations. In natural language processing, techniques might encompass synonym replacement or back-translation. 

In related works, one early study~\cite{sablayrolles_EjVuEmpirical_2018} demonstrates trivial data augmentation can reduce the risks of membership inference attacks, thereby diminishing memorization. Utilizing recent memorization evaluation methods, Li et al.~\cite{li_PrivacyEffectData_2023} study the memorization effect of multiple data augmentation. They measure the memorization evaluation results by membership inference and demonstrate trivial data augmentation technologies significantly mitigate memorization. However, for advanced data augmentation technologies, further research on the memorization effect is still required. Another work~\cite{anagnostidis_CuriousCaseBenign_2022} measures memorization based on $k$NN probing and they find that $k$NN probing accuracy of the embedding vectors increases with data augmentation under random label training datasets and clean training datasets. Moreover, they observe that learning under complete label noise with data augmentation still leads to highly useful features in the shallow layers, explaining it as augmented datasets increasing the effective size of the dataset beyond the capacity of networks. This supports that data augmentation mitigates memorization. However, the mechanism of how data augmentation impacts memorization still needs to be explored and systematically evaluated.

\paragraph{Regularization}
Regularizers like weight decay and dropout are the standard tools in theory and practice to mitigate overfitting in the training of neural networks. We know that regularizers help constrain the learning process to a specific subset of the hypothesis space with manageable complexity. In the work of Zhang et al.~\cite{zhang_UnderstandingDeepLearning_2017}, explicit regularizers can prevent model memorization under random label learning, and help the model improve generalization. However, regularization is neither necessary nor by itself sufficient for controlling generalization errors. Then the research conducted by Arpit et al.~\cite{arpit_CloserLookMemorization_2017} reproduces a similar result as Zhang et al.~\cite{zhang_UnderstandingDeepLearning_2017} and finds dropout is best at hindering memorization without reducing the model's ability to learn. This also responds to the work of location memorization~\cite{maini_CanNeuralNetwork_2023}, in which finds memorization exists in a small set of neurons in various layers of the model. It seems under random label training, explicit regularizers can mitigate memorization by dropping or constraining neurons, but it is not clear to understand how regularizers influence atypical example memorization in standard training.

\textbf{Summary.} Data augmentation and regularization are standard tools in training neural networks and help improve model generalization. In related works, both tools mitigate memorization under random label training, but we do not know if they hinder learning long-tailed examples in standard training. This could be a research direction in the future.

\subsection{Memorization and Other Factors}
\paragraph{Capacity} 
The model capacity is related to model memorization. Generally, models with larger sizes can memorize more data than smaller ones~\cite{carlini_QuantifyingMemorizationNeural_2023}. Additionally, early work has shown that overparameterized neural networks can directly memorize randomly labeled modern datasets~\cite{zhang_UnderstandingDeepLearning_2017}. However, we cannot easily think larger capacity leads to more memorization because training data plays an important role. The effective capacity of networks cannot directly explain memorization and generalization~\cite{arpit_CloserLookMemorization_2017}. Naturally, a question arises: what happens if the training dataset size far exceeds the model's capacity? Data augmentation can create this condition, and Anagnostidis et al.~\cite{anagnostidis_CuriousCaseBenign_2022} find that even the randomly labeled dataset with data augmentation exceeding the model capacity can produce effective patterns in the model. Nevertheless, related topics still require further research.

\paragraph{Loss} 
The loss function is an important component of neural network training. Thus, it must influence the memorization dynamics of models. Patel et al.~\cite{patel_MemorizationDeepNeural_2021} propose robust log loss (RLL) which can prevent model overfitting on the randomly labeled data. However, no further studies have explored how different types of loss functions affect model memorization. 

\paragraph{Learning Rate}
Learning rate is an essential hyperparameter in neural network training. Li et al.~\cite{li_ExplainingRegularizationEffect_2019} believe that a small learning rate model easily learns details, while a large learning rate helps capture patterns. They demonstrate this by adding a small patch to CIFAR10 images that are immediately memorizable by a model with a small initial learning rate but ignored by the model with a large learning rate until after annealing.

\paragraph{Data Format}
The data format may affect memorization during training, particularly for language tasks. Kharitonov et al.~\cite{kharitonov_HowBPEAffects_2021} find the size of the subword vocabulary learned by Byte-Pair Encoding (BPE) greatly affects both ability and tendency of standard Transformer models to memorize training data. Larger subword vocabulary and shorter input sequences result in strong memorization of Transformer models. The underlying reason could be that complex subwords weaken the patterns in the data distribution. Thus, the input data format likewise impacts memorization. 

\textbf{Summary.} Many other factors may facilitate or hinder memorization. Specifically, large models tend to memorize due to an excessive number of parameters. Moreover, a small learning rate can promote memorization. Additionally, loss functions and data format have impacts on model memorization but we still require further studies.

\section{Underlying Risks of Memorization Learning}~\label{sec:securityAndPrivacy}
\begin{figure*}[!htbp]
    \centering
    \includegraphics[width=1\textwidth]{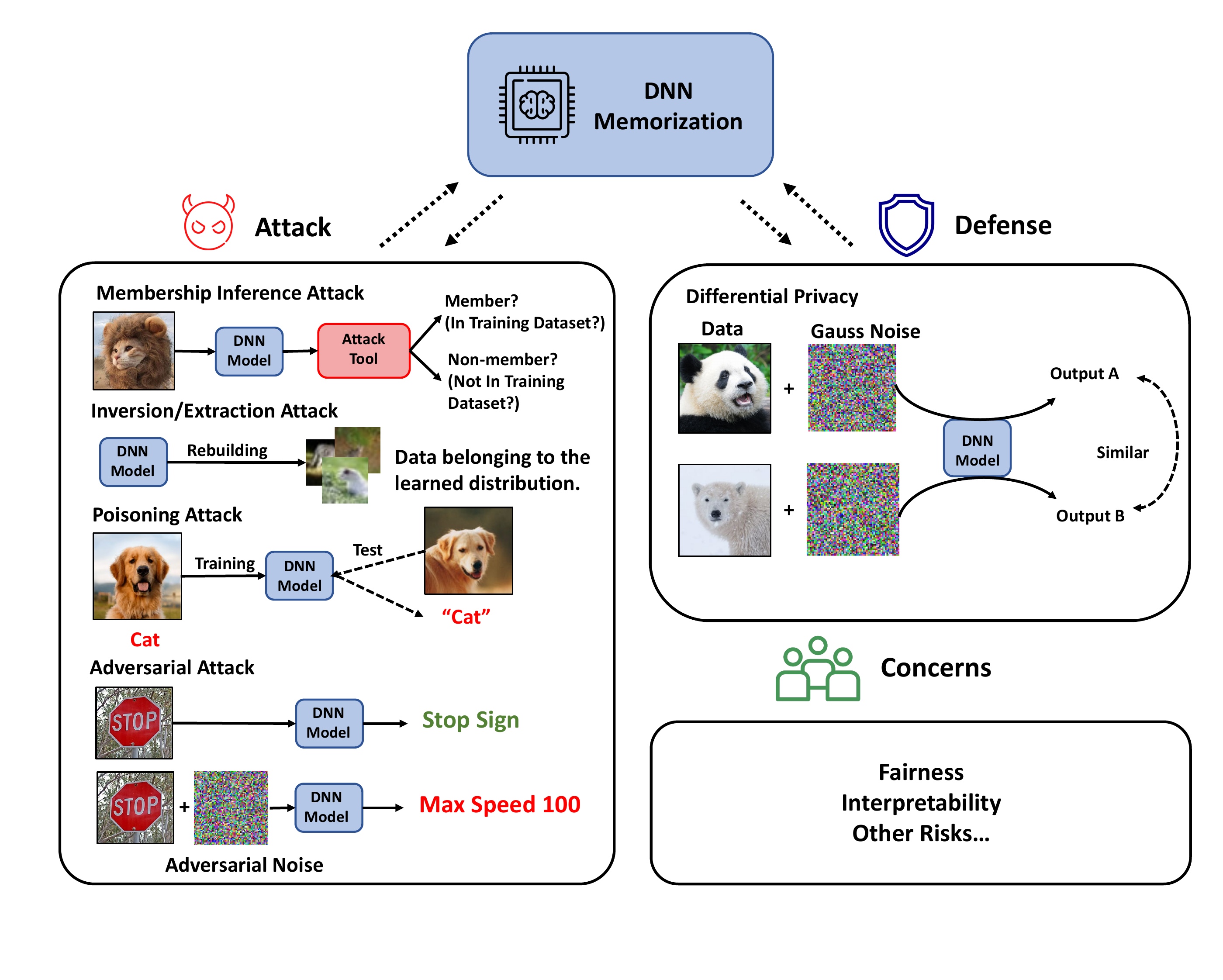}
    \caption{Underlying Risks of Memorization.}
    \label{fig:risks}
\end{figure*}

In previous sections, DNNs have been shown the feature of memorizing training data, and this property may cause various security risks. This section undertakes exploration and synthesis of the impact of memorization on typical threats and defenses in DNNs. We summarize the main literature related to the risks of memorization learning in Table~\ref{table:risksMemorizationLearning} and plot Figure~\ref{fig:risks} to demonstrate these risks.

\begin{table*}[!htbp]
\centering
\scriptsize
\caption{Main Works about Underlying Risks of Memorization Learning}
\label{table:risksMemorizationLearning}
%\fontsize{4}{4}\selectfont
%\resizebox{1\linewidth}{!}{
\begin{tblr}{
  cells = {c,m},
  colspec = {X[1.2] X[1] X[1.2] X[1.5] X[3]},
  cell{2}{1} = {r=3}{},
  cell{5}{1} = {r=3}{},
  cell{8}{1} = {r=3}{},
  cell{11}{1} = {r=3}{},
  hline{1-2,5,8,11,14} = {-}{1pt},
  hline{-} = {-}{},
}
\textbf{Main Topic}                       & \textbf{Reference}                          & \textbf{Background Task}                       & \textbf{Main Eval. Method}                                  & \textbf{Main Findings}                                                                                                                     \\
Memorization about
  Membership Inference Risks & Leino et al., 2020~\cite{leino_StolenMemoriesLeveraging_2020}                & Supervised Classification Task        & /                                     & Propose a new membership inference attack based on memorized features.                                                                     \\
                                                & Carlini et al., 2022~\cite{carlini_MembershipInferenceAttacks_2022}  & Various Tasks                         & /                                     & Propose a new membership inference attack 'LiRA' utilizing the
  memorization effect.                                                      \\
                                                & Carlini et al., 2022~\cite{carlini_PrivacyOnionEffect_2022}       & Supervised Classification Task        & Probabilistic Memorization Evaluation & Removing the vulnerable outlier points may threaten inner previously-safe
  points on the same attack.                                     \\
Memorization about
  Extraction Risks           & Carlini et al., 2019~\cite{carlini_SecretSharerEvaluating_2019}      & Unsupervised Language Generative Task & Recurrence Memorization Evaluation    & Introduce a memorization exposure metric to measure unintended
  memorization.                                                             \\
                                                & Carlini et al., 2021~\cite{carlini_ExtractingTrainingData_2021}  & Unsupervised Language Generative Task & Extraction Memorization Evaluation    & An adversary can perform a training data extraction attack to recover
  individual training examples by querying the large language model. \\
                                                & Carlini et al., 2023~\cite{carlini_QuantifyingMemorizationNeural_2023} & Unsupervised Language Generative Task & Extraction Memorization Evaluation    & Describe three log-linear relationships that how model capacity,
  duplication times, and the number of tokens impact memorization of LMs.~ \\
Memorization about
  Poisoning Risks            & Zhang et al., 2017~\cite{zhang_UnderstandingDeepLearning_2017}                 & Supervised Classification Task        & Noisy Label Memorization Evaluation   & DNNs can memorize randomly labeled datasets that traditional approaches
  fail to explain generalization.~                                  \\
                                                & Nguyen et al., 2023~\cite{nguyen_MemorizationDilationModelingNeural_2023}                & Supervised Classification Task        & Noisy Label Memorization Evaluation   & Mislabeled examples may degrade the neural collapse and damage model
  generalization                                                      \\
                                                & Maini et al., 2023~\cite{maini_CanNeuralNetwork_2023}                & Supervised Classification Task        & Noisy Label Memorization Evaluation   & Drop memorization-associated neurons, mislabeled examples cannot be
  effectively classified.                                              \\
Memorization about
  Adversarial Risks          & Li et al., 2023~\cite{li_PrivacyEffectData_2023}                    & Supervised Classification Task        & Probabilistic Memorization Evaluation & In adversarial training, adversarial examples are very atypical and prone
  to be memorized.                                               \\
                                                & Xu et al., 2023~\cite{xu_HowDoesMemorization_2023}                    & Supervised Classification Task        & Differential Memorization Evaluation  & Memorizing atypical samples hardly improve their adversarial robustness.                                                                   \\
                                                & Dong et al., 2022~\cite{dong_ExploringMemorizationAdversarial_2022}                  & Supervised Classification Task        & Noisy Label Memorization Evaluation   & Memorization in adversarial training could result in robust overfitting.                                                                   
\end{tblr}
\end{table*}

\subsection{Memorization and Membership Inference Risks}

A membership inference attack is a representative privacy inference attack and seeks to address the query if a specific instance belongs to the training dataset~\cite{shokriMembershipInferenceAttacks2017}. In the machine learning setting, the membership inference adversary is typically given access to a model’s predictions with varying granularity, ranging from the complete confidence vector to the label corresponding to the highest confidence score. It is established that the memorization phenomenon entails the memorization of training data points by DNNs, thereby implying that memorized data bears substantial risks for membership inference.

Indeed, prior work has demonstrated that the membership inference risks associated with training data exhibit significant non-uniformity. According to empirical results, typical data points have lower membership inference risks than those atypical data examples and outliers~\cite{carlini_MembershipInferenceAttacks_2022,feldman_WhatNeuralNetworks_2020,leino_StolenMemoriesLeveraging_2020,jagielski_MeasuringForgettingMemorized_2022}. These findings imply memorization is highly corresponding or even results in high membership inference risks. However, no direct quantitative investigation has been identified to establish the precise relationship between memorization and membership inference risks. In practice, this relationship has been implicitly approved~\cite{carlini_PrivacyOnionEffect_2022,carlini_MembershipInferenceAttacks_2022,leino_StolenMemoriesLeveraging_2020}. 

Depending on the relationship between memorization and membership inference attack, Carlini et al.~\cite{carlini_PrivacyOnionEffect_2022} find the privacy onion effect. The effect can be defined: when removing the most vulnerable data under a specific privacy attack and retraining a model on only the previously safe data, a new set of examples in turn becomes vulnerable to the same privacy attack. This phenomenon may indicate that even after removing those memorized data points, the model still memorizes relatively atypical data examples in the remaining training dataset. This observation intuitively underscores membership inference risks are highly associated with memorization and proves that memorization is relative.

Utilizing the memorization effect, researchers investigate more threatening membership inference attacks. Leino et al.~\cite{leino_StolenMemoriesLeveraging_2020} attempt to exploit features of memorized examples that are predictive only for the training data but not the sampling distribution. They capture differences in memorization learning data and pattern learning data and build a confident binary logistic classifier to infer membership.  Another work is Likelihood Ratio Attack (LiRA)~\cite{carlini_MembershipInferenceAttacks_2022}, it depends on the leave-one-out method as memorization score definition and utilizes differences of model outputs that training with and without the training example to do membership inference. These attacks directly exhibit that memorized examples have higher privacy risks than generalized examples.

\textbf{Summary.} While no direct quantitative research has yet proven the relationship between memorization and membership inference attacks, nearly all relevant existing studies suggest a strong correspondence between memorization and membership inference risks and even indicate a causal relationship. Additionally, it is imperative to explore novel inference risks and mitigation strategies that are contingent upon the memorization effect.

\subsection{Memorization and Inversion/Extraction Risks}

The adversary in an inversion/extraction attack attempts to rebuild or extract training examples by leveraging gradients or models. The attack obviously threatens the privacy of machine learning as the acquired training examples inherently unveil sensitive information. Based on current knowledge, the generalized features embedded in the gradient or model parameters cannot facilitate the precise reconstruction or extraction of training examples because these features are common. Consequently, the underlying reasons for the inversion/extraction risk potentially come from the memorization phenomenon.

Related work mainly analyses the memorization effect concerning the extraction risk of language tasks~\cite{carlini_SecretSharerEvaluating_2019,carlini_ExtractingTrainingData_2021,carlini_QuantifyingMemorizationNeural_2023}. Carlini et al.~\cite{carlini_SecretSharerEvaluating_2019} introduce a memorization exposure metric utilizing canary sequences and log-perplexity. Subsequently, they establish that successful extraction becomes feasible when the level of memorization exposure surpasses a threshold. Conversely, extraction remains unsuccessful below this threshold. Consequently, it can be inferred that memorized examples carry a substantial risk of extraction.

Another work~\cite{carlini_ExtractingTrainingData_2021} directly demonstrates the performance of their proposed extraction attack applied to GPT-2. The researchers generate an extensive dataset through unconditional sampling from the model and employ diverse metrics to identify examples exhibiting high memorization likelihood. Consequently, they find the extraction result actually consists of trivial memorization and atypical information. We can explain the outcome originating from two distinct inversion/extraction mechanisms. The first mechanism rebuilds representations of common knowledge based on patterns. Another mechanism extracts atypical and individual examples exactly depending on memorized data. If we consider the two through the lens of privacy, the former mechanism supports the main task of DNNs while the latter apparently breaks privacy. 

Moreover, if we acknowledge the robust correlation between memorization and the inversion/extraction risk, the measurement outcomes of inversion/extraction risk can be regarded as an empirical lower-bound of memorization~\cite{carlini_QuantifyingMemorizationNeural_2023}.   

\textbf{Summary.} Previous studies indicate the vulnerability of inversion/extraction risks is highly relevant to the memorization effect within models. Despite the absence of direct experimental evidence, we attempt to reveal two inversion/extraction mechanisms: one where the attack reconstructs representations of common knowledge through generalized patterns, and another where it precisely extracts exceptional and individual examples using memorized data. Consequently, we deduce that the memorization effect constitutes the foundational cause of privacy risk in inversion/extraction attacks. Furthermore, the outcomes of such attacks can serve as a lower-bound approximation of model memorization.

%Add Cite
\subsection{Memorization and Poisoning Risks}
Poisoning attacks target breaking model availability. Specifically, adversaries attempt to degrade model performance on all examples (i.e. \textit{untargeted poisoning attack}) or specific classes or examples (i.e. \textit{targeted poisoning attack}), even examples with particular features (i.e. \textit{backdoor attack}). Common poisoning techniques include data manipulation, called \textit{data poisoning}, and model corruption, called \textit{model poisoning}. As \textit{model poisoning} is generally used in distribution machine learning systems, we mainly discuss data poisoning including label manipulation and input noise corruption.

Randomly labeled examples cannot be classified to the assigned noisy class based on pattern learning due to the absence of highly shared features. However, based on empirical results~\cite{zhang_UnderstandingDeepLearning_2017}, we know that DNNs can minimize the training loss of randomly labeled datasets and achieve almost perfect accuracy. Therefore, randomly labeled examples are memorized and we can infer that data poisoning via label manipulation depends on the memorization effect. Arpit et al.~\cite{arpit_CloserLookMemorization_2017} provides more effective evidence that model performance reduction is quantitatively corresponding to the proportion of mislabeled examples. Additionally, a recent study~\cite{maini_CanNeuralNetwork_2023} finds if drop memorization-associated neurons, mislabeled examples cannot be effectively classified. However, they also demonstrate atypical examples and noisy examples are hard to identify for DNNs and generalization may be damaged by dropping memorization-associated neurons. Memorization-dilation~\cite{nguyen_MemorizationDilationModelingNeural_2023} based on neural collapse provides an explanation that mislabeled examples may degrade the neural collapse and damage model generalization. 

Another technology of data poisoning is adding random noise to the input. As noise increases, we can infer the inputs may gradually transform from typical examples to atypical examples then full noise. Concurrently, DNNs have been forced to minimize the loss so models may boost the memorization of such inputs.

\textbf{Summary.} When an adversary launches a poisoning attack with mislabeled data, neural networks would memorize these data to minimize the loss. Thus, memorization is an adaptive process and the vulnerability to poisoning attack comes from the DNNs training framework.

\subsection{Memorization and Adversarial Risks}
The adversarial attack employs adversarial noise on inputs to drive examples approaching the decision boundary and achieving the maximum loss. Generally, the adversarial noise is generated by gradient ascending~\cite{madry_DeepLearningModels_2017, goodfellow_ExplainingHarnessingAdversarial_2014}. An effective defense strategy is adversarial training~\cite{madry_DeepLearningModels_2017} which means directly training on the adversarial examples and this method provides a lower-bound robustness guarantee. 

In spite of the absence of relevant studies on memorization and adversarial attacks, we can infer that the memorization effect is not the source of adversarial vulnerability because existing work~\cite{ilyas_AdversarialExamplesAre_2019} believes adversarial vulnerability derives from non-robust features.

Many works~\cite{xu_HowDoesMemorization_2023,li_PrivacyEffectData_2023,dong_ExploringMemorizationAdversarial_2022} investigate memorization in adversarial training. Li et al.~\cite{li_PrivacyEffectData_2023} find adversarial examples are very atypical and Schmidt et al.~\cite{schmidt_AdversariallyRobustGeneralization_2018} demonstrate the complexity of adversarial examples can be significantly larger than normal examples in standard learning, thus DNNs memorize them during adversarial training making DNNs more vulnerable to privacy attacks. Another work~\cite{xu_HowDoesMemorization_2023} exhibits that memorizing atypical examples hardly helps adversarial robustness and when memorizing some harmful atypical examples that share similar features with a “wrong" class, the boundary becomes blurred, and this damages robustness. This phenomenon may be explained by robust overfitting that one-hot labels can be inappropriate and some adversarial examples should be given low confidence~\cite{rice_OverfittingAdversariallyRobust_2020}. Researchers~\cite{dong_ExploringMemorizationAdversarial_2022} are also curious about randomly labeled dataset performance in adversarial training and they find PGD-AT~\cite{madry_DeepLearningModels_2017} fails to converge while TRADES~\cite{zhang_TheoreticallyPrincipledTradeoff_2019a} still reaches nearly 100\% training accuracy. However, they believe DNNs have sufficient capacity to memorize adversarial examples of training data with completely random labels, but the convergence depends on the AT algorithms. Next, they analyze the gradients of the two different adversarial training methods and recognize the gradient of PGD-AT performs large variance, making it fail to converge. Moreover, they study robust overfitting and put it down to excessive memorization of one-hot labels breaking the robust decision boundary.

\textbf{Summary.} It seems that the memorization effect is not responsible for adversarial vulnerability. In the adversarial training, due to the hardness of examples, most adversarial examples will be memorized increasing privacy leakage. Particularly, memorizing examples sharing similar features with a “wrong" class or excessive memorization of one-hot labels may corrupt the decision boundary.

\subsection{Memorization and Differential Privacy}
Differential Privacy~\cite{abadi_DeepLearningDifferential_2016} (DP) is a commonly employed strategy for defending against privacy attacks, which aims to guarantee indistinguishability between various data points. In particular, DP ensures that the trained model remains largely unchanged when any single example is removed from the training set.  Within the framework of $(\epsilon,\delta)$-DP setting, DP comprises two key components: gradient clipping and the application of noise. Gradient clipping restricts the gradients of each example to a predefined boundary, reducing disparities in their gradient magnitudes. This facilitates the standardization of gradients in terms of magnitude and mitigates the memorization effect. The phenomenon has been observed in some cases~\cite{thakkar_UnderstandingUnintendedMemorization_2020, carlini_MembershipInferenceAttacks_2022}. Additionally, the random noise application also promotes example memorization reduction. When models are trained on examples mixed with random noise, the features carried by long-tailed examples or atypical features will be diluted, leading the models to mainly learn typical patterns. Moreover, neural networks may memorize artificial random noise~\cite{feldman_DoesLearningRequire_2020,feldman_WhatNeuralNetworks_2020,carlini_ExtractingTrainingData_2021}. Because we always observe that effective DP measures hurt model generalization. Corresponding, from a privacy standpoint, we can understand that DP operates by safeguarding privacy through the prevention of atypical feature memorization. While a comprehensive analysis of DP from a memory perspective is currently lacking, some researchers agree that DP can limit example memorization supported by empirical results~\cite{carlini_MembershipInferenceAttacks_2022,leino_StolenMemoriesLeveraging_2020,carlini_SecretSharerEvaluating_2019}.

\textbf{Summary.} DP serves as an effective measure to alleviate the issue of example memorization in neural networks, achieved through gradient clipping and the introduction of noise. Nevertheless, the efficacy of DP may be guaranteed by reducing the memorization of atypical features and increasing the learning of random noise.

\subsection{Memorization and Other Risks}
As DNNs have been widely applied in various social scenarios, the public gradually shifted its focus from system performance to additional attributes such as model fairness~\cite{caton2020fairness, dwork2012fairness, feldman2015certifying, hardt2016equality, 
zemel2013learning}, interpretability~\cite{selvaraju2017grad, simonyan2013deep, fong2017interpretable, dwivedi2023explainable, guidotti2018survey}, and others. These additional attributes potentially exhibit a strong relationship with the phenomenon of memorization, consequently leading to additional risks. In terms of fairness, DNNs may inadvertently learn societal biases present in the training data from the real world~\cite{tramer2017fairtest}. Such biases come from unbalanced subgroups or sensitive attributes, which correlates with the long-tailed theory~\cite{feldman_DoesLearningRequire_2020}. This indicates the significant role of memorization in the risk of unfairness. Furthermore, certain approaches~\cite{gao_FairNeuronImprovingDeep_2022,barone_IncreasingFairnessClassification_2024} aimed at promoting fairness employ techniques like data augmentation and weight reassignment, which may amplify privacy leakage via memorization learning. Regarding interpretability, the public views uninterpretable models and predictions as uncontrolled risks. Memorization study can help mitigate this risk. Additionally, the phenomenon of memorization is also associated with certain risks of violation of intellectual property rights or copyrights.

\textbf{Summary.} We believe that the memorization effect has strong relationships with fairness, interpretability, and other risks. However, this area of research remains largely unexplored.

\section{Forgetting Research}~\label{sec:forgetting}
\begin{figure*}[!htbp]
    \centering
    \includegraphics[width=0.8\textwidth]{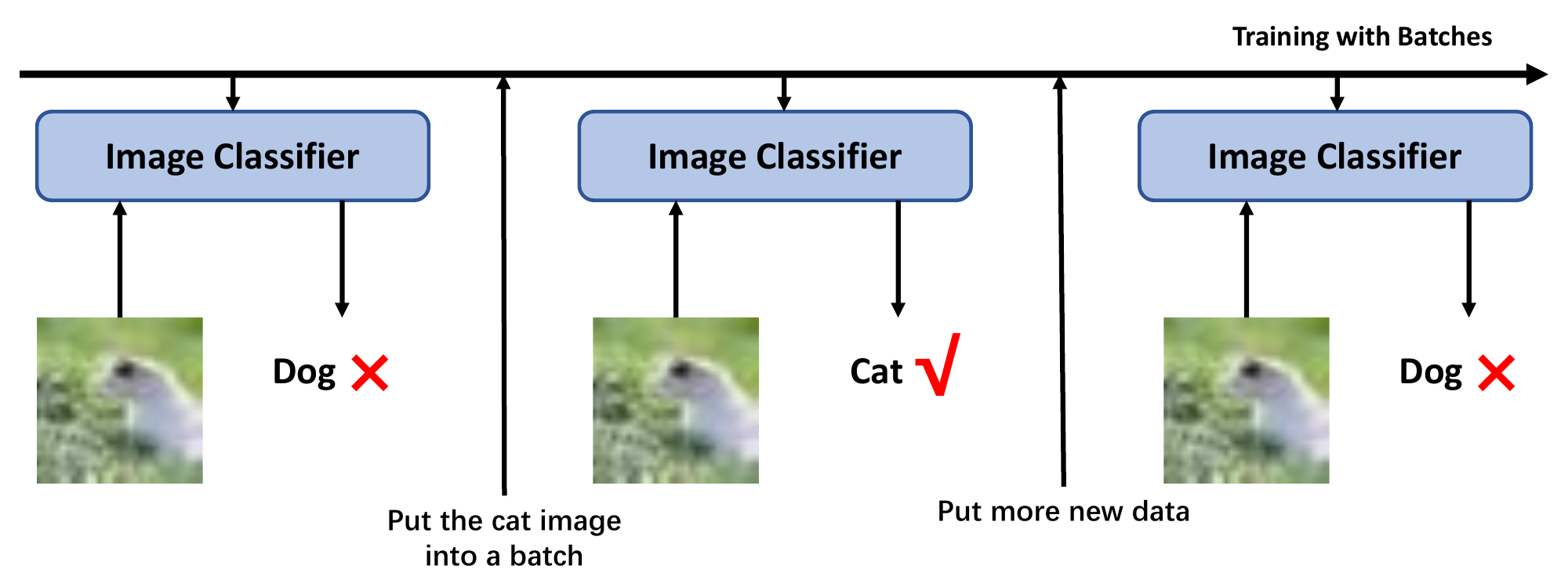}
    \caption{Demonstration of the Forgetting Phenomenon.}
    \label{fig:forgettingFigure}
\end{figure*}

Forgetting is the opposite of memorization. Generally, neural networks may encounter difficulties in continual learning because the learning capacity of networks is not infinite. During iterative training, as networks train on new examples, they tend to forget learned features or information from previous examples, as shown in Figure~\ref{fig:forgettingFigure}. This phenomenon is known as catastrophic forgetting~\cite{ritter_OnlineStructuredLaplace_2018,kirkpatrick_OvercomingCatastrophicForgetting_2017}. Variations in data distributions cause models to converge to different optimal points. Although there are some methods to overcome this phenomenon~\cite{shao_OvercomingCatastrophicForgetting_2022,kirkpatrick_OvercomingCatastrophicForgetting_2017,ritter_OnlineStructuredLaplace_2018,hayes_REMINDYourNeural_2020,liu_OvercomingCatastrophicForgetting_2021,li_LearningForgetting_2018,shao_OvercomingCatastrophicForgetting_2022}, we still lack an understanding of forgetting especially as an opposite of memorization. We may be curious about what information will be forgotten, how the forgetting effect impacts model performance and privacy, and its relationship with memorization. The main works are presented in Table~\ref{table:forgetting}.

\begin{table*}
\centering
\scriptsize
\caption{Main Works about Forgetting}
\label{table:forgetting}
\begin{tblr}{
  cells = {c,m},
  colspec = {X[1.2] X[1] X[1.3] X[1.5] X[3]},
  cell{2}{1} = {r=2}{},
  hline{1,2,4-5} = {-}{1pt},
  hline{-} = {-}{},
}
\textbf{Main Topic}                       & \textbf{Reference}                          & \textbf{Background Task}                       & \textbf{Main Eval. Method}                                  & \textbf{Main Findings}     \\
Forgetting about
  Data  & Toneva et al. 2019~\cite{toneva_EmpiricalStudyExample_2019}    & Supervised Classification Task & Forgetting and Learning Event         & Atypical examples and noisy examples are prone to be forgotten.                                            \\
                         & Maini et al. 2022~\cite{maini_CharacterizingDatapointsSecondSplit_2022}     & Supervised Classification Task & Second-Split Forgetting Time          & In fine-tune, noisy examples are forgotten quickly and seemingly atypical
  examples are forgotten slowly. \\
Forgetting about Privacy & Jagielski et al. 2022~\cite{jagielski_MeasuringForgettingMemorized_2022} & Various Tasks                  & Probabilistic Memorization Evaluation & Standard image, speech, and language models empirically do forget
  examples over time.                    
\end{tblr}
\end{table*}

\subsection{Forgetting Definition and Evaluation}
% Existing studies~\cite{toneva_EmpiricalStudyExample_2019,maini_CharacterizingDatapointsSecondSplit_2022, jagielski_MeasuringForgettingMemorized_2022} include multiple definitions of example forgetting. Here we combine these definitions:
% \begin{definition}~\label{forgetting}
% \textbf{example Forgetting Definition.} Given a model $f$ and an example $i$, designing a metric $M$ and a forgetting comparison formula $C$, the model $f$ forgets the example $i$ at step $t^*$ comparing to $t$ which can be expressed as:
% \begin{align}
% C(M(f,i,t),M(f,i,t^*))
% \end{align}
% \end{definition}

\subsubsection{Forgetting Definition based on Accuracy}

Toneva et al.~\cite{toneva_EmpiricalStudyExample_2019} propose the forgetting and learning event:
\begin{definition}~\label{def:forgetEvent} 
\textbf{Forgetting and Learning Event.} For supervised classification task, given an example $\mathbf{x}_i$, the predicted label for example $\mathbf{x}_i$ obtained after $t$ steps of SGD is $\hat{y}_i^t = \arg \max_k p(y_{ik}|\mathbf{x}_i;\theta^t)$ and accuracy is $acc_i^t = \vmathbb{1}_{\hat{y}^t_i = y_i}$. Therefore, the forgetting event is that example $i$ is misclassified at step $t + 1$ after having been correctly classified at step $t$ (i.e. $acc^t_i > acc^{t+1}_i$). Conversely, a learning event has occurred if $acc^t_i < acc^{t+1}_i$.
\end{definition}

Following the forgetting and learning event definitions, Maini et al.~\cite{maini_CharacterizingDatapointsSecondSplit_2022} definite First-Split Learning Time (FSLT) to demonstrate the first epoch that model learns an example and Second-Split Forgetting Time (SSFT) to describe the forgetting time in the fine-tuned stage.
\begin{definition}~\label{def:FSLT}
\textbf{First-Split Learning Time.} For $\{\mathbf{x_i}, y_i\}\in D_A$, learning time is defined as the earliest epoch during the training of a classifier $f$ on $D_A$ after which it is always classified correctly, i.e.
    \begin{IEEEeqnarray}{rCl}
        FSLT_i = \arg\min_{t^*}(\hat{y}_{i,(A)}^t = y_i, \forall t \geq t^*) \ \forall\{\mathbf{x_i}, y_i\} \in D_A,
    \end{IEEEeqnarray}
where $t$ denotes epoch, $A$ is the pre-training stage and $D_A$ is the training dataset. The $f_A$ represents the trained model with 100\% training accuracy on the $D_A$. 
\end{definition}

\begin{definition}~\label{def:SSFT}
\textbf{Second-Split Forgetting Time.} Let $\hat{y}_{i,(A \rightarrow B)}^t$ to denote the prediction of example $\{\mathbf{x_i}, y_i\}\in D_A$ after training $f_{(A \rightarrow B)}$ for $t$ epochs on $D_B$. Then, for $\{\mathbf{x_i}, y_i\}\in D_A$ forgetting time is defined as the earliest epoch after which it is never classified correctly, i.e.,
    \begin{IEEEeqnarray}{rCl}
        SSFT_i = \arg\min_{t^*}(\hat{y}_{i,(A \rightarrow B)}^t \neq y_i, \forall t \geq t^*) \ \forall\{\mathbf{x_i}, y_i\} \in D_A,
    \end{IEEEeqnarray}
where $D_B$ is a held-out split dataset (without $\{\mathbf{x_i}, y_i\}$) of $D_A$, $f_{(A \rightarrow B)}$ is initialized by $f_{A}$.
\end{definition}

These definitions can be used to measure forgetting in supervised classification tasks. Based on the nature of forgetting, i.e. learned features have been lost, it is reasonable to obverse forgetting depending on the accuracy.

\subsubsection{Forgetting Definition based on Membership Inference Attack}
Jagielski et al.~\cite{jagielski_MeasuringForgettingMemorized_2022} measure the ratio of forgetting based on a membership inference attack.

\begin{definition}~\label{def:forgetMIA} 
\textbf{Rate of Forgetting.} A training algorithm $\mathcal{T}$ is said to $(\mathcal{A}, \alpha, k)$-forget a training example $z$ if, $k$ steps after $z$ is last used in $\mathcal{T}$, a privacy attack $\mathcal{A}$ achieves no higher than success rate $\alpha$.
\end{definition}

We know that the membership inference attack relies on particular features, thus, the reduced risk could be regarded as forgetting. This is also an effective definition to describe forgetting.

\subsection{Forgetting Phenomenon}
Some existing studies provide interesting evidence to understand the forgetting phenomenon. Toneva et al.~\cite{toneva_EmpiricalStudyExample_2019} study example forgetting during DNN learning and use accuracy as the metric to define the example forgetting event, i.e. an example that has been correctly classified becomes misclassified. They find that generalized examples are unforgettable (always correctly classified) but atypical examples and noisy examples are prone to be forgotten. This is connected with the related memorization findings~\cite{feldman_DoesLearningRequire_2020,feldman_WhatNeuralNetworks_2020} that forgetting occurs on those memorized examples. Even with some intermediate forgetting events, DNNs still can correctly classify all training examples and finally achieve 100\% training accuracy, indicating memorization is a challenging but forced learning stage. Another finding is removing a part of these generalized/unforgettable examples will not damage the model generalization performance. This may be explained as DNNs do not need to repeatedly learn common patterns. 

Following this research, Maini et al.~\cite{maini_CharacterizingDatapointsSecondSplit_2022} propose second-split forgetting time (SSFT) to track the epoch (if any) after which an original training example is forgotten as the network is fine-tuned on a randomly held-out partition of the data. In the fine-tuned stage, they demonstrate that noisy examples are forgotten quickly and seemingly atypical examples are forgotten slowly, while typical examples are never forgotten. Tirumala et al.~\cite{tirumala_MemorizationOverfittingAnalyzing_2022} employ Definition~\ref{def:exactMem} to measure the single-injected validation dataset forgetting dynamics and find the exact memorization from a higher point gradually drops to the forgetting baseline as the number of epoch increases. The forgetting baseline may represent generalization. 

Jagielski et al.~\cite{jagielski_MeasuringForgettingMemorized_2022} focus on the privacy risk associated with forgetting. They utilize the membership inference probability to evaluate forgetting and believe that the size of the training dataset, repetitions, and hardness mainly influence forgetting. Examples used early in model training may be more robust to privacy attacks, while repeated examples are harder to forget. During the forgetting phase, the membership inference probability of typical examples is still around 50\% which is lower than the inference risk of atypical examples. This finding corresponds to previous studies. They also illustrate that non-convexity and deterministic SGD can prevent forgetting. Another piece of evidence is the variation in local data distribution of every batch of data leads optimization techniques to converge to different local optimal points. Therefore, some lack-of-representativeness information may be lost.

\textbf{Summary.} In the single-task learning scenario, memorized examples based on the view of performance metrics are easy to forget. However, the low performance does not mean all features of these memorized examples have been forgotten because they still pose high inference risks compared to generalized examples.

% \textbf{Summary.} Based on these studies, in the single-task learning scenario, we can assume that the root reason for forgetting is the variation in local data distribution leads optimization techniques to converge to different points. Moreover, we can infer that the generalized features and common patterns of the entire data distribution are unforgettable because later-learned examples also utilize these features, while some memorized particular features may perhaps be replaced or forgotten. Repeated features have been learned more times. Thus, they are difficult to forget.

\section{Application}~\label{sec:application}
Utilizing the memorization and forgetting effects of neural networks, researchers have developed various applications in several scenarios. We organize these applications in Tabel~\ref{table:application}.

\begin{table*}[!htbp]
\centering
\scriptsize
\caption{Related Application about Memorization and Forgetting}
\label{table:application}
\begin{tblr}{
  cells = {c,m},
  colspec = {X[1] X[2] X[1.5] X[2]},
  cell{2}{1} = {r=3}{},
  cell{2}{2} = {r=2}{},
  cell{6}{1} = {r=3}{},
  cell{6}{2} = {r=2}{},
  cell{9}{1} = {r=2}{},
  hline{1-2,5-6,9,11} = {-}{1pt},
  hline{-} = {-}{}
}
\textbf{Application}                   & \textbf{Associated Memorization or
  Forgetting Effect}                                   & \textbf{Technology}                                  & \textbf{Reference}                                                                                         \\
Noisy Label Learning           & DNNs are prior to learning patterns in the early training
  stage.                      & Noise Control Scheduler                     & Han et al., 2018~\cite{han_CoteachingRobustTraining_2018}; Yao et al., 2020~\cite{yao_SearchingExploitMemorization_2020}                                                                  \\
                               &                                                                                  & Regularization                              & Liu et al., 2020~\cite{liu_EarlyLearningRegularizationPrevents_2020}; Xia et al., 2020~\cite{xia_RobustEarlylearningHindering_2020}                                                                   \\
                               & Pre-training on random labels
  can learn label-irrelevant generalized features. & Random Labels Pre-training                  & Pondenkandath et al., 2018~\cite{pondenkandath_LeveragingRandomLabel_2018}; Maennel et al., 2020~\cite{maennel_WhatNeuralNetworks_2020}                                                     \\
Example Enhancement            & Long-tailed examples are prone to be memorized.                                  & Example Reweighting                         & Zhou et al., 2022~\cite{zhou_ContrastiveLearningBoosted_2022}; Xu et al., 2023~\cite{xu_HowDoesMemorization_2023};
  Zhang et al., 2023~\cite{zhang_MemorizationWeightsInstance_2023}                                              \\
Privacy Audit and
  Protection & Memorization cause inference and extraction risks                               & Membership Inference and Extraction Attacks & Carlini et al., 2021~\cite{carlini_ExtractingTrainingData_2021}; Carlini et al., 2022~\cite{carlini_MembershipInferenceAttacks_2022}                                 \\
                               &                                                                                  & Memorization Suppression                    & Maini et al., 2023~\cite{maini_CanNeuralNetwork_2023}                                                                                 \\
                               & The privacy of forgotten
  examples are relatively guaranteed.                  & Unlearning Technology                       & Zhu et al., 2020~\cite{zhu_ModifyingMemoriesTransformer_2020}                                                                                   \\
Memorization
  Architecture    & Explicit memorization helps specific task performance.                           & Explicit Memorization Structure             & Khandelwal et al., 2019~\cite{khandelwal_GeneralizationMemorizationNearest_2019}; Yogatama et al., 2021~\cite{yogatama_AdaptiveSemiparametricLanguage_2021}; Guu et al., 2020~\cite{guu_RetrievalAugmentedLanguage_2020}; Lewis et al., 2020~\cite{lewis_PretrainingParaphrasing_2020}; Lewis et
  al., 2020~\cite{lewis_RetrievalAugmentedGenerationKnowledgeIntensive_2020}; Wu et al., 2022~\cite{wu_MemorizingTransformers_2022} \\
                               & DNNs could be viewed as
  databases or knowledge-bases                           & DNN Database                                & Tay et al., 2022~\cite{tay_TransformerMemoryDifferentiable_2022a}
  \\
   Model Editing	& Language models memorize a lot of facts.	& Memorization Neurons Modification	& Dai et al., 2021~\cite{dai2021knowledge}; De Cao et al., 2021~\cite{de2021editing}; Mitchell et al., 2021~\cite{mitchell2021fast}; Meng et al., 2022~\cite{meng2022locating,meng2022mass}; Gupta et al., 2024~\cite{gupta_UnifiedFrameworkModel_2024}.
\end{tblr}
\end{table*}

\subsection{Noisy Label Learning}
Deep learning with noisy labels is challenging, as neural networks have powerful memorization abilities that can completely memorize all noisy labels in the later training stage. However, this memorization phenomenon is utilizable. 

Specifically, one effective method for learning with noisy labels involves selecting potentially clean learning examples in each iteration for training~\cite{yu_HowDoesDisagreementHelp_2019,han_CoteachingRobustTraining_2018a,jiang_MentorNetLearningDataDriven_2018,yao_SearchingExploitMemorization_2020}. In example selection, it is challenging to choose a reasonable threshold based on example loss to drop underlying noisy examples because dropping too many examples may lose some useful features and lead to lower accuracy~\cite{han_CoteachingRobustTraining_2018a}. According to previous studies~\cite{arpit_CloserLookMemorization_2017}, we know that DNNs usually learn easy patterns before overfitting the noisy examples, and the pattern learning period can be regarded as the early learning stage. Thus, in the early stage of noisy label learning, it is not necessary to drop a large number of examples and the threshold could be loose. As training epochs increase, the dropping rate will also increase to avoid noisy label memorization. Therefore, it requires to define a scheduler used to control the example selection. Han et al.~\cite{han_CoteachingRobustTraining_2018a} propose a novel pre-defined scheduler. The scheduler is a non-increasing function and is controlled by noise level and current epoch. At the start of training, the scheduler would not drop any examples, but as the training continues, the dropping rate would increase. 
% Han et al.~\cite{han_CoteachingRobustTraining_2018a} give the scheduler as 
% $R(t) = 1- \tau \cdot \min((t/t_k)^c,1).$ In the formula, $t$ represents the number of epoch, $\tau$ depends on the noise level $\epsilon$, $t_k$ implies the noise memorization epoch, and $c$ denotes a scaling hyperparameter. The $R(t)$ actually satisfies the following requirements: i). $R(t) \in [\tau,1]$, where $\tau$ depends on the noise rate $\epsilon$; ii). $R(1) = 1$, which means no dropping in the first epoch; iii) $R(t)$ is a non-increasing function on $t$, which means that avoiding noise memorization requires dropping more instances when learning proceeds. 
However, this pre-defined scheduler may not be "optimal" and is limited to specific tasks and datasets. Yao et al.~\cite{yao_SearchingExploitMemorization_2020} improve the scheduler as an AutoML problem to conduct a search, which outperforms previous works.

Loss modification and regularization technology represent another approach to learning with noisy labels. Existing works~\cite{liu_EarlyLearningRegularizationPrevents_2020,xia_RobustEarlylearningHindering_2020} attempt to control noise during the early-learning stage. Liu et al.~\cite{liu_EarlyLearningRegularizationPrevents_2020} develop early-learning regularization (ELR). This regularization item can facilitate learning from clean examples by prioritizing pattern learning and restraining noise in the training dataset by maximizing the inner product between the model output and the targets. %The entire loss is $\mathcal{L}_{ELR}(\theta)=\mathcal{L}_{CE}(\theta) + \frac{\lambda}{n}\sum\limits_{i=1}^n \log(1-<\mathbf{p}^{[i]},\mathbf{v}^{[i]}>)$, where $\theta$ is the model parameters, $\mathbf{p}^{[i]}$ denotes the output probability vector of example $i$ and $\mathbf{v}^{[i]}_t = \beta \mathbf{v}^{[i]}_{k-1} + (1-\beta)\mathbf{p}^{[i]}_k$. 
Specifically, the regularization item can diminish noisy examples' effect on the gradient, implicitly preventing the memorization of wrong labels. Modifying the gradient update based on the differences between pattern-associated neurons and memorization-associated neurons can provide a similar regularization effect. Xia et al.~\cite{xia_RobustEarlylearningHindering_2020} find parameters have different tendencies that some respond for fitting clean labels as critical parameters and some for memorization as non-critical parameters.  It is possible to assess the parameter tendency in each iteration based on the inner product between the value of the parameters and the gradient with respect to the parameters. Then, performing positive normal gradient updates on critical parameters and applying weight decay only to non-critical parameters mitigates noise learning during the early learning stage. If the minimum classification error is achieved on the validation dataset, the training should be stopped early.

Another way to utilize the memorization effect in noisy label learning is pre-training, expecting that models can learn label-irrelevant useful features. Basically, the data distribution always has some low-level and label-irrelevant features. For instance, in colorful image classification, the data distribution contains colorful low-level features compared to the gray in random noise. One previous study~\cite{pondenkandath_LeveragingRandomLabel_2018} empirically proves that performing unsupervised pre-training in the form of training for classification with random labels may boost initialized training speed and improve the generalization performance. However, these label-irrelevant features are all low-level generalized features and can be easily learned in the normal training process. Maennel et al.~\cite{maennel_WhatNeuralNetworks_2020} demonstrate that the pre-training on random labels may pose a positive effect or negative effect on downstream tasks. They show that aligned shallow layers improve the performance of downstream tasks. However, the neural activations at the deep layers may drop abruptly and permanently on downstream tasks due to specialization, which may impair downstream task performance.

\subsection{Example Enhancement}
Data distribution is not uniform in the real environment. Atypical examples with low frequencies in the data distribution always exist and cause models to memorize them instead of learning patterns based on the long tail theory~\cite{feldman_WhatNeuralNetworks_2020,feldman_DoesLearningRequire_2020}. Some existing research applies example reweighting technology to handle these long-tailed examples. 

In self-supervised long-tailed representation learning, learning long-tailed examples is generally challenging. Zhou et al.~\cite{zhou_ContrastiveLearningBoosted_2022} employ the memorization effect to boost the performance of contrastive learning on tail examples. Specifically, they attempt to identify tail examples and apply heavier augmentation, consistently improving the performance of these examples. Due to the high computational cost associated with tracking memorization using memorization scores~\cite{feldman_WhatNeuralNetworks_2020}. They tend to trace the historical losses of each example as memorization clues. Then, they construct the normalization of momentum losses which indicates the memorization level of examples. Based on the memorization level, a stronger information discrepancy between views will be constructed to emphasize the importance of tail examples. Their results demonstrate that the method is effective.

For adversarial training, Xu et al.~\cite{xu_HowDoesMemorization_2023} discover that memorizing atypical examples is only effective in improving DNN’s accuracy on clean atypical examples but hardly improves their adversarial robustness. Moreover, fitting some atypical adversarial examples even damages the model's robustness. They believe some atypical examples share similar features with a wrong class and become harmful in the context of adversarial training. This may uncover the key differences between traditional standard training and adversarial training. Motivated by their findings, they propose an algorithm called Benign Adversarial Training (BAT), which can mitigate the negative influence of memorizing those harmful atypical examples, simultaneously preserving the model’s ability to learn those useful/benign atypical examples. In BAT, the core is the example reweighting technology that assigns the harmful atypical examples a small weight to reduce the negative effect. Additionally, harmful atypical examples can be detected by the high memorization score and high misclassification confidence in the standard training. Another related work~\cite{zhang_MemorizationWeightsInstance_2023} also agrees the atypical examples may hurt DNN's robustness. They similarly employ example reweighting technology to reassign example update weights. Their method actually builds a $k$NN structure to measure the example typicalness. This structure has been called the codebook and trained concurrently with the classifier. The codebook training is actually clustering based on distance, finding the central points approaching the nearest embedding vectors. These central points could be regarded as the most typical embedding vectors. Then, depending on the distance between input batch embedding and central vectors in the codebook, it is effective in identifying the atypical examples. Subsequently, the distances can be utilized as weights to enhance atypical example learning in standard training and reduce atypical example learning in adversarial training.

\subsection{Privacy Audit and Protection}
With a deeper understanding of the memorization phenomenon, some works~\cite{thakkar_UnderstandingUnintendedMemorization_2020,carlini_ExtractingTrainingData_2021,carlini_MembershipInferenceAttacks_2022,carlini_QuantifyingMemorizationNeural_2023,carlini_SecretSharerEvaluating_2019} have demonstrated that memorized particular features become sources of risk for several attacks. Thus, researchers can apply the memorization effects to audit security risks and develop novel defense strategies. 

In section~\ref{sec:securityAndPrivacy}, we have introduced memorization-associated risks. For membership inference attacks, it is possible to utilize the memorized particular features of examples to achieve high true-positive rates at low false-positive rates~\cite{carlini_MembershipInferenceAttacks_2022}. Moreover, extractable training examples also should depend on those particular features~\cite{carlini_ExtractingTrainingData_2021}. However, adversely, we actually can employ these attacks as memorization audit tools to help model compliance. 

Additionally, mitigating the memorization effect may reduce the corresponding risks. Maini et al.~\cite{maini_CanNeuralNetwork_2023} propose a novel dropout technology. This technology utilizes the memorization effect to guide the specific neurons to memorize the specific details of examples, while other neurons work for pattern learning. When the neural network drops these memorization neurons, those memorized particular features are dropped along with the neurons. Another promising strategy is applying the forgetting mechanism to unlearning technology. Private features of long-tailed examples could be forgotten under continuous learning with tail-changing data distribution. This means the new data distribution after removing the target unlearning example could not provide similar features for the target unlearning example, and features of the example will be gradually forgotten because parameters related to the example may be updated. Zhu et al.~\cite{zhu_ModifyingMemoriesTransformer_2020} apply the forgetting phenomenon without retraining to update a Transformer on the new integrated dataset without stale knowledge. However, it is still unsure if we can control the forgetting process and achieve the expected unlearning effect.

\subsection{Memorization Architecture}
Although memorization leads to privacy risks, we may require the memorization ability of networks. Specifically, memorizing or caching common input and output can improve inference speed. Networks or large language models (LLMs) also can function as an information retrieval system. Despite this, some external memory structures like $k$NN and key-value memory could serve as additional components to improve task performance.

Khandelwal et al.~\cite{khandelwal_GeneralizationMemorizationNearest_2019} attempt to combine the $k$NN and language model. They find that applying $k$NN as an external memorization structure can improve performance. Qualitatively, this kind of model is particularly helpful in predicting rare patterns that allow rare features to be memorized explicitly rather than implicitly in model parameters. Search based on similarity is better than predicting the next word in the long tail. Moreover, they also extend this method to the machine translation~\cite{khandelwal_NearestNeighborMachine_2020} and achieve non-trivial performance. Yogatama et al.~\cite{yogatama_AdaptiveSemiparametricLanguage_2021} modify this approach by a gating mechanism and context compression retrieval.

Additionally, the explicit key-value memory can also improve the effectiveness of inference. In the context of Transformers, F\text{\'e}vry et al.~\cite{fevry_EntitiesExpertsSparse_2020} and Verga et al.~\cite{verga_FactsExpertsAdaptable_2020} apply similar methods to install an external key-value memory to store entities and facts. With this explicit memorization, models can achieve higher performance in question-answering tasks. 

For information retrieval, REALM~\cite{guu_RetrievalAugmentedLanguage_2020}, MARGE~\cite{lewis_PretrainingParaphrasing_2020}, and RAG~\cite{lewis_RetrievalAugmentedGenerationKnowledgeIntensive_2020} apply knowledge retrieval in pre-training. REALM~\cite{guu_RetrievalAugmentedLanguage_2020} augments language model pretraining with a knowledge retriever, which allows the model to retrieve documents from a large corpus, used during pre-training, fine-tuning, and inference. Next, MARGE~\cite{lewis_PretrainingParaphrasing_2020} is trained by self-supervising the reconstruction of the target text. This process first involves retrieving a set of related texts (in many languages) and then maximizing their likelihood of generating the original. RAG~\cite{lewis_RetrievalAugmentedGenerationKnowledgeIntensive_2020} combines parametric memory and non-memory for language generation. The parametric memory is a pre-trained seq2seq model and the non-parametric memory is a dense vector index of Wikipedia. These pre-training models all gain non-trivial results in the downstream tasks like question-answering tasks. Regarding the latest studies, Wu et al.~\cite{wu_MemorizingTransformers_2022} envision language models that can simply read and memorize new data at inference time, thus acquiring new knowledge immediately. By using attention, a model can simply memorize facts (e.g. function definitions) by storing them as key-value pairs in long-term memory. Then, it can retrieve those facts later by creating a query that attends to them. In this case, attention acts as a form of information retrieval, allowing the model to look up facts that it has seen previously. Thus, they propose a Transformer model with $k$NN-augmented attention that unifies attention and retrieval. Their experiment demonstrates that an approximate $k$NN lookup into a non-differentiable memory of recent key-value pairs improves language modeling across various benchmarks and tasks. 

A more interesting work directly regards Transformer memory as a differentiable search index (DSI)~\cite{tay_TransformerMemoryDifferentiable_2022a}. All information about the corpus is encoded in the parameters of a Transformer. In other words, a DSI model answers queries directly using only its parameters, dramatically simplifying the whole retrieval process. At inference time, the trained model takes as input a text query and outputs the id of the correlated document.

\subsection{Model Editing}
Due to the computational burden of training large language models (LLMs) and the requirement of updating information in LLMs, researchers attempt to directly edit LLMs neurons to update facts~\cite{dai2021knowledge, de2021editing, mitchell2021fast, meng2022locating, meng2022mass,
gupta_UnifiedFrameworkModel_2024}. LLMs learn a variety of facts about the world during pre-training and these facts are stored in model weights~\cite{petroni2019language}. Specifically, the MLP weights actually serve as key-value memories~\cite{geva2020transformer, geva2022transformer, meng2022locating}. Thus, editing these neurons to update facts becomes a practical approach.

Dai et al.~\cite{dai2021knowledge} first identify knowledge-containing neurons in a model using integrated gradients~\cite{sundararajan2017axiomatic} and then modify the selected neurons to edit facts in a model. Specifically, they focus on evaluating BERT's performance on the fill-in-the-blank cloze task. In this task, they introduce a technique called knowledge attribution, aiming to find the neurons in BERT that represent specific facts. Their analysis demonstrates a positive correlation between the activation of these identified 'knowledge neurons' and the accurate expression of the corresponding facts. De Cao et al.~\cite{de2021editing} and Mitchell et al.~\cite{mitchell2021fast} train a hypernetwork that predicts the new weights of the model being edited. This method modifies a fact rapidly without affecting the rest of the knowledge.

Meng et al.~\cite{meng2022locating,meng2022mass} develop two "locating and editing" technologies: Rank-One Model Editing (ROME) and Mass-Editing Memory In a Transformer (MEMIT). They create a method for causal intervention to identify the activation of neurons crucial for a model's factual predictions. Then, they directly update particular "knowledge-containing" components of the model without requiring to train additional models. Additionally, this approach is applicable to any transformer-based LLM. Subsequently, Gupta et al.~\cite{gupta_UnifiedFrameworkModel_2024} build a unifying conceptual framework for ROME and MEMIT following the preservation-memorization objective of model editing. During the editing process, this approach preserves the representations of certain selected vectors while memorizing the representations of new factual information. In summary, model editing demonstrates a good prospect of memorization utilization.

\section{Discussion and Future Research}~\label{sec:discussion}

The memorization effect of DNN is an ongoing field with significant implications for the interpretability, generalization, and security of AI. In this section, we will discuss existing research findings and possible future research directions.

\subsection{Memorization and Forgetting Mechanism}
The memorization and forgetting mechanism remains unclear and confusing. However, based on existing studies, we have known some memorization and forgetting truths on the classification task: 
\begin{itemize}
    \item Standard training framework always leads DNNs to the minimal loss~\cite{papyan2020prevalence};
    \item DNNs can memorize common modern training datasets, even when the dataset is randomly labeled~\cite{zhang_UnderstandingDeepLearning_2017};
    \item Long-tailed examples that lack representation in the data distribution like atypical examples and noisy examples are prone to be memorized~\cite{feldman_DoesLearningRequire_2020,feldman_WhatNeuralNetworks_2020};
    \item DNNs cannot identify atypical examples and noisy examples in training~\cite{maini_CanNeuralNetwork_2023};
    \item DNNs tend to prioritize the memorization of repeated data~\cite{zhang_CounterfactualMemorizationNeural_2021,lee_DeduplicatingTrainingData_2022,carlini_QuantifyingMemorizationNeural_2023};
    \item DNNs have a critical early learning stage where pattern learning takes domination~\cite{arpit_CloserLookMemorization_2017};
    \item Memorization appears to be confined to a limited set of neurons across various layers in DNNs~\cite{maini_CanNeuralNetwork_2023};
    \item Memorization is not responsible for overfitting~\cite{long_UnderstandingMembershipInferences_2018,yeom_PrivacyRiskMachine_2018,maini_CanNeuralNetwork_2023};
    \item Long-tailed examples are prioritized to be forgotten, and noisy examples are forgotten more quickly than atypical examples~\cite{maini_CharacterizingDatapointsSecondSplit_2022,toneva_EmpiricalStudyExample_2019}.
\end{itemize}

According to these observations, it is reasonable to infer that, at least in the context of classification tasks, the memorization phenomenon is a property of the standard DNN gradient descent training framework. Specifically, minimizing the loss leads to the memorization effect. For instance, when a neural network has been fed some data during training, some parameters are activated and updated. The network may gradually build several mapping paths of common patterns, and each input example passes these mapping paths and then gains a probability score for every class. These probability scores constitute a probability vector and the predicted label will be the class with the maximum score. However, long-tailed examples cannot gain reasonable scores on all of these mapping paths because they have no strong patterns compared to more generalized examples. Consequently, the network may build a unique mapping path for each long-tailed example to minimize the loss, and this is the memorization phenomenon. 

Considering training with the batch stochastic gradient descent method, in the beginning, the network may randomly create unique mapping paths for each example to fit the labels. Subsequently, because of inherent patterns of data distribution, some of these mapping paths gradually align. The alignment can be regarded as the pattern extraction and may correlate with the early learning phenomenon~\cite{achille2018critical,frankle_EarlyPhaseNeural_2019,arpit_CloserLookMemorization_2017} because the alignment effect decreases the loss of representative examples in the data distribution. This alignment precedes memorization because early gradients represent the direction that can most effectively reduce the loss. Additionally, long-tailed examples may also participate in the alignment but not align well, which let these examples experience forgetting events~\cite{toneva_EmpiricalStudyExample_2019} (Definion~\ref{def:forgetEvent}), i.e. some long-tailed examples could be correctly classified at previous steps but misclassified at later steps.  Simultaneously, the network could employ some extra capacity or parameters to memorize some particular features of long-tailed examples that are not aligned well in the early learning stage to reduce the loss. This is evidenced by concurrent improvements in accuracy for randomly labeled examples and clean examples~\cite{maini_CanNeuralNetwork_2023}. This also explains why memorization does not depend on overfitting. Moreover, this indicates memorization learning and pattern learning are not totally discrete and contrary. They imply the difficulty of pattern extraction on examples. The more challenging the pattern extraction, the more apparent the tendency to be memorized. After the alignment phase, the network will prioritize memorizing long-tailed examples, which can lead to close-to-optimal generalization error based on the long tail theory~\cite{feldman_DoesLearningRequire_2020,feldman_WhatNeuralNetworks_2020}. After memorizing of long-tailed examples, the network may exhibit the best generalization performance. However, if the training continues, diminishing the loss becomes challenging. The network may develop unique paths for all examples, causing the predicted vector to closely approach the label vector, even resulting in zero training error. This phenomenon is referred to as neural collapse~\cite{papyan2020prevalence}, where each example in the same class collapses to the same representation. For architecture, unique memorization mapping paths may require only a few parameters across layers because these paths are not based on pattern recognition. Therefore, it is reasonable to observe that memorization appears to be confined to a limited set of neurons across various layers in DNNs~\cite{maini_CanNeuralNetwork_2023}. Furthermore, if we take into account forgetting, these memorized examples become highly unstable. This instability arises because even if a small part of the associated parameters has been updated, these examples are likely to be misclassified. This phenomenon explains why long-tailed examples are particularly prone to being forgotten~\cite{maini_CharacterizingDatapointsSecondSplit_2022}. Meanwhile, there may remain some unchanged associated parameters that pose privacy risks~\cite{jagielski_MeasuringForgettingMemorized_2022}.

Certainly, our theoretic model of memorization and forgetting in the classification task is an assumption. Further experimentation and empirical evidence are required to fully explain the memorization phenomenon. Understanding the memorization mechanism carries significant implications for enhancing the interpretability of DNNs. To effectively understand this mechanism, it is crucial to describe the spatiotemporal memorization process. In terms of training periods, the primary objectives include characterizing memorization in different stages and investigating whether the memorization phenomenon constitutes a form of overlearning. Concerning neural network components, it becomes essential to quantitatively explain the distribution of memorization across layers or components and evaluate whether certain neurons exhibit a tendency for memorizing examples. Additionally, it is important to differentiate between memorization learning and pattern learning neurons. Moreover, different training frameworks may have distinct memorization phenomena, particularly for unsupervised tasks and multiple-task scenarios.

\subsection{Memorization and Forgetting for Training Discussion}
\paragraph{Data}
DNNs offer significant advantages in processing complex real-world data such as images and text compared to traditional machine learning methods~\cite{krizhevsky_ImageNetClassificationDeep_2012,vaswani_AttentionAllYou_2017a}. A notable observation is that DNNs can effectively extract common features or patterns from the data distribution. However, controlling this extraction process is challenging, and some uncommon yet useful features may not be learned well~\cite{feldman_DoesLearningRequire_2020,feldman_WhatNeuralNetworks_2020}. At the feature level, out-of-distribution features and rare but useful features are both less representative. This may explain why memorization learning cannot identify atypical examples and noisy examples. Moreover, we may rethink how to describe complex data in reality to keep features balanced. This encourages us to contemplate aspects such as data dimension, granularity~\cite{cui_MeasuringDatasetGranularity_2019}, and distribution~\cite{chu_FeatureSpaceAugmentation_2020,zhang_DistributionAlignmentUnified_2021} to enhance the performance of DNNs. Additionally, the size of the training dataset probably does not serve as the sole determining factor for task performance~\cite{althnian_ImpactDatasetSize_2021}.

\paragraph{Training Framework}
It is understood that the stochastic gradient descent method will aggressively minimize the loss function until reaching extreme mathematical conditions such as neural collapse~\cite{papyan2020prevalence}. However, the extreme conditions may not meet our requirements, and even potentially introduce further challenges. From this perspective, the loss function really matters and decides the learning direction. The challenge lies in the fact that loss functions may not always accurately measure the true loss associated with the assigned task. For instance, in a classification task, a model with minimal loss may have poor generalization performance due to overfitting. Therefore, the memorization and forgetting effect may serve as adaptive solutions to address this conflict.

\paragraph{Architecture}

The impact of neural network architecture details on the memorization phenomenon remains unclear. Different layers within the architecture may assume distinct roles, with certain layers potentially exhibiting a preference for memorization. Viewing the architecture in terms of layer depth, deeper layers may tend to learn more specialized features~\cite{yosinski_HowTransferableAre_2014} although these features are not completely for memorization. Specialized features still retain patterns, whereas memorized features may lack patterns and serve primarily to mark data. Therefore, deeper layers do not function for memorization~\cite{maini_CanNeuralNetwork_2023}. Regarding the size of networks, the memorization tendency also correlates with the size of the training dataset. A larger model trained on a small dataset may lead to significant overfitting and a strong inclination toward memorization. Conversely, larger datasets, which contain more diverse patterns, tend to reduce the preference for memorization but may increase the probability of underfitting.    

\paragraph{Tasks}
Memorization and forgetting manifest differently across various tasks. Presently, most memorization studies focus on the classification task, where the memorization phenomenon entails the utilization of a small set of parameters to uniquely mark examples. The classification task, being a dimension reduction task, can apply this way to minimize the loss. However, for other tasks such as generative tasks, the dynamics differ. Obviously, the generative task could be a dimension increment task like GAN~\cite{goodfellow2020generative}, Diffusion model~\cite{dhariwal_DiffusionModelsBeat_2021}, and GPT~\cite{radford_ImprovingLanguageUnderstanding_}, which aim to learn a target data distribution. Thus, the generalization of generative models refers to the model's ability to produce accurate, relevant, and coherent outputs to cover the target data distribution. From this viewpoint, the memorization phenomenon in the generative task could be significantly different from the classification task. For instance, generative models may use enormous parameters to memorize almost all features of those long-tailed examples. This phenomenon has been demonstrated in some works~\cite{carlini_ExtractingTrainingData_2021,carlini_ExtractingTrainingData_2023}. Furthermore, multiple-task learning and continuous learning are also distinct. A famous phenomenon called catastrophic forgetting~\cite{kirkpatrick_OvercomingCatastrophicForgetting_2017,shao_OvercomingCatastrophicForgetting_2022,ritter_OnlineStructuredLaplace_2018} means that neural networks are hard to retain learned knowledge on multiple and dynamic data distributions. As the data distribution shifts based on sub-tasks, and the model learning capacity is limited, separating learned features becomes challenging. In such scenarios, memorization may have a very beneficial effect in preserving learned features.

\subsection{Memorization and Forgetting for Privacy and Security Discussion}
\paragraph{Privacy Leakage}
Privacy leakage is a common problem in neural networks. Direct privacy leakage includes the inference attack and the inversion/extraction attack we mentioned before. According to some empirical evidence~\cite{carlini_MembershipInferenceAttacks_2022,jagielski_MeasuringForgettingMemorized_2022}, the vulnerability of privacy leakage often arises from memorization. For membership inference attacks, unique features of examples are heavily exploited rather than representative features that generally improve false-positive rates~\cite{carlini_MembershipInferenceAttacks_2022}. It is known that examples sharing patterns but not in the training set are very easily inferred as membership. This protects the privacy of those representative examples. Conversely, memorized long-tailed examples are more vulnerable. In inversion/extraction attacks, the adversary can generally produce representative data based on patterns. However, the representative data could be common knowledge and not private. The real and particular data belonging to the training dataset is more valuable for the adversary. Due to varied behaviors of memorization in various tasks, generative models are more vulnerable from the perspective of memorization because the generative models may be required to memorize more long-tailed examples to fit the target data distribution. The memorization process likely contains most features of certain examples, allowing these examples to be reconstructed in a lossless manner under inversion/extraction attacks. Some empirical results have illustrated that memorization is the source of extraction~\cite{carlini_ExtractingTrainingData_2023,carlini_ExtractingTrainingData_2021}. Qualitatively, the memorization effect indeed poses risks of privacy leakage. Therefore, mitigating the memorization effect may reduce risks associated with inference and inversion attacks. Nonetheless, we acknowledge that memorization somehow contributes to task performance~\cite{feldman_DoesLearningRequire_2020,feldman_WhatNeuralNetworks_2020}. Considering a trade-off framework could be beneficial.

\paragraph{Malicious Attacks}
Poisoning, backdoor, and adversarial attacks are typical malicious attacks. These attacks can directly disable networks or embed malicious triggers to mislead models. Due to the absence of specific threat evaluation, we only conduct some hypothetical discussions. Suppose the attack just modifies the training data without optimization and targets disabling networks like label flip and random noise. In such cases, the model fails because it cannot learn correct patterns following the gradient direction and has to memorize them. Therefore, the memorization phenomenon is an adaptive process. For induced attacks without optimization~\cite{chen_TargetedBackdoorAttacks_2017}, these attacks can install malicious triggers in networks, the triggers could be learned via pattern learning or memorization learning. Specifically, this depends on the feature distribution of triggers. If the backdoor feature is long-tailed, here applying memorization, otherwise it is pattern learning. Finally, some malicious attacks are based on optimization, the adversary can submit artificial features or gradients to the networks~\cite{madry_DeepLearningModels_2017}. The synthetic features or gradients are out of the standard training framework, so it is challenging to discuss the memorization effect under this condition. This requires further studies.

\paragraph{Forgetting Guarantee}
From the perspective of privacy, the forgetting phenomenon actually provides the privacy guarantee by ensuring that previously learned particular features of examples will be forgotten in later stages of training without prompt repetition~\cite{tirumala_MemorizationOverfittingAnalyzing_2022}. Therefore, the input order of examples may also impact the example privacy. Forgetting also corresponds to machine unlearning technology. At the example level, typical examples may only provide generalized patterns. Unlearning these examples is not valid because other examples also provide similar features. However, for long-tailed examples, after removing these examples from the training dataset, the model will gradually forget particular features of them during constant training and the corresponding privacy risk is simultaneously reducing~\cite{jagielski_MeasuringForgettingMemorized_2022}. However, the privacy onion effect~\cite{carlini_PrivacyOnionEffect_2022} is another problem. This effect indicates that the removal of some most vulnerable data could improve the threat of other vulnerable data. Consequently, it is uncertain how the forgetting phenomenon and relevant unlearning technologies quantitatively reduce privacy risks.

\paragraph{Memorization Inhibition}
Considering the underlying privacy risks of memorization, we could apply some technologies or strategies to inhibit memorization. In relevant technologies, data augmentation and regularization can help inhibit memorization and improve generalization. Data augmentation enhances the patterns within the data, particularly for those long-tailed examples. Regarding regularizers, weight decay technology restrains the feature space and prevents extreme parameters used in memorization, while the dropout strategy could randomly drop memorization activation. Moreover, it's feasible to guide memorization to specific neurons and drop them when testing~\cite{maini_CanNeuralNetwork_2023}. However, this technology may drop some features of atypical examples. Despite this, we can also follow the memorization mechanism to propose some new regularizers to mitigate the memorization effect. Differential privacy is another effective tool. The added noise would cover some features that tend to be memorized, potentially causing the networks to memorize the added noise rather than unique features. However, this may result in performance degradation. Thus, while memorization inhibition can protect privacy, it may simultaneously impair generalization performance.

\paragraph{Threat Evaluation}
The perspective of memorization and forgetting offers a novel and interesting entry point to re-evaluate existing threat and defense strategies. The memorization and forgetting mechanism can enhance the interpretability of certain threats, thereby deeply understanding existing threats. From this perspective, we could propose new solutions or enhance existing defense methods. Additionally, this viewpoint can assist in uncovering previously unknown threats. It is essential to systematically assess how memorization and forgetting influence these threats and defense methods.

\subsection{Application Discussion}
As memorization is an innate feature of DNNs, we can apply the advantages of the memorization effect to assist in some specific tasks and mitigate the disadvantages in some scenarios. 

In positive terms, large language models directly benefit from memorization abilities in tasks like question-answer. Memorization can serve as an additional structure to cache representations~\cite{khandelwal_GeneralizationMemorizationNearest_2019,khandelwal_NearestNeighborMachine_2020} or key-value pairs, improving network speed and performance~\cite{fevry_EntitiesExpertsSparse_2020,verga_FactsExpertsAdaptable_2020}. Furthermore, networks can function as databases or knowledge bases with the memorization effect~\cite{tay_TransformerMemoryDifferentiable_2022a}. Additionally, the memorization of LLMs can be modified directly to update facts. LLMs learn a variety of facts about the world during pretraining and these facts are stored in model weights~\cite{petroni2019language}, suggesting that MLP weights act as key-value memories~\cite{geva2020transformer, geva2022transformer, meng2022locating}. Thus, editing memorization neurons for injecting new information has been a popular technology called model editing to update facts which sidesteps the computational burden associated with training a wholly new model~\cite{sinitsin2020editable, de2021editing, dai2021knowledge, meng2022locating, meng2022mass, mitchell2021fast, mitchell2022memory, gupta_RebuildingROMEResolving_2024, hase_DoesLocalizationInform_2024, gupta_RebuildingROMEResolving_2024, yao_EditingLargeLanguage_2023}.

Indirectly, the memorization phenomenon can be employed to filter noisy examples and atypical examples, which benefits example enhancement~\cite{zhou_ContrastiveLearningBoosted_2022,xu_HowDoesMemorization_2023,zhang_MemorizationWeightsInstance_2023} and noisy data learning~\cite{han_CoteachingRobustTraining_2018,yao_SearchingExploitMemorization_2020,liu_EarlyLearningRegularizationPrevents_2020,xia_RobustEarlylearningHindering_2020,pondenkandath_LeveragingRandomLabel_2018,maennel_WhatNeuralNetworks_2020}.

However, the memorization effect also raises privacy concerns. We may utilize the memorization effect to audit privacy~\cite{carlini_QuantifyingMemorizationNeural_2023,carlini_ExtractingTrainingData_2021} or mitigate memorization to ensure compliance~\cite{maini_CanNeuralNetwork_2023}. Additionally, the forgetting effect may provide some privacy guarantee and correlate with unlearning technology.

\section{Conclusion}

This survey based on the memorization effect provides a detailed exploration of a pivotal concept in DNNs. 
It begins by discussing the memorization definitions in the generalization domain and security and privacy domain. 
The survey then provides relevant measurements of memorization at different levels. 
Next, we discuss how memorization influences DNN training including data distribution, training stage, model structure, and other factors. 
After that, we review related studies on underlying privacy and security risks that correlate with the memorization effect.
Following this, 
we also review the studies about the forgetting effect because forgetting is the opposite of memorization. 
Subsequently, 
this survey discusses related applications to the memorization effect or are highly associated with memorization functions.
Lastly, 
we discuss possible memorization and forgetting mechanisms,
attempt to understand memorization and forgetting impacts, 
and suggest further research. 
In this survey, 
to the furthest extent, 
we collect and present the main literature about the memorization and forgetting effect of DNNs, 
organizing relevant works in a comprehensive framework. 

In this review, 
we highlight that memorization and forgetting effects are features of DNNs. 
These effects have deep impacts on the performance, fairness, explainability, accountability, and privacy of DNNs. 
Therefore, we should develop the ability to control, manage, and utilize the effects, leading to highly usable and trustworthy neural networks.

\bibliographystyle{IEEEtran}
\bibliography{reference}

%\input{Chapters/bio}

% Biography text here without a photo.

% \begin{IEEEbiography}[{\includegraphics[width=1in,height=1.25in,clip,keepaspectratio]{fig1.png}}]{IEEE Publications Technology Team}
% In this paragraph you can place your educational, professional background and research and other interests.\end{IEEEbiography}

\end{document}